\documentclass[10pt, conference, compsocconf]{IEEEtran}
\ifCLASSINFOpdf
\else
\fi

\usepackage[pdftex]{graphicx}
\usepackage{comment}
\usepackage{url}
\usepackage{xcolor}
\usepackage{soul}
\usepackage [autostyle, english = american]{csquotes}
\usepackage{enumitem}
\usepackage{booktabs}
\usepackage{multicol}
\usepackage{multirow}
\usepackage{verbatim}
\usepackage[linesnumbered]{algorithm2e}
\usepackage{algorithmicx}
\usepackage{algpseudocode}
\usepackage{caption}
\usepackage{amsmath}
\usepackage{amssymb}
\usepackage{amsfonts}
\usepackage{amstext}
\usepackage{cleveref}

\usepackage{amsthm}
\usepackage{diagbox}
\usepackage{makecell}
\DeclareCaptionType{copyrightbox}

\theoremstyle{definition}

\newcounter{myexample}
\newenvironment{myexample}[1][]{\refstepcounter{myexample}\par
   \indent \textbf{Example~\themyexample. #1} \rmfamily}%{\smallskip}

\usepackage{subcaption}
\MakeOuterQuote{"}

\begin{document}
\title{Boosting Entity Mention Detection for Targetted Twitter Streams with Global Contextual Embeddings}

\author{\IEEEauthorblockN{Sataisha Saha Bhowmick}
\IEEEauthorblockA{Department of Computer Science\\
Binghamton University\\
%Binghamton, New York\\
Email: ssahabh1@binghamton.edu
}
\and
\IEEEauthorblockN{Eduard C. Dragut}
\IEEEauthorblockA{Department of Computer Science\\
Temple University\\
%Temple, Philadelphia\\
Email: edragut@temple.edu}
\and
\IEEEauthorblockN{Weiyi Meng}
\IEEEauthorblockA{Department of Computer Science\\
Binghamton University\\
%Binghamton, New York\\
Email: meng@binghamton.edu}}


%Telephone: (800) 555--1212, Fax: (888) 555--1212}


\maketitle
\thispagestyle{plain} %--to make title page number
\pagestyle{plain}    % -- to make other pages number

\begin{abstract}
Microblogging sites, like Twitter, have emerged as ubiquitous sources of information. Two important tasks related to the automatic extraction and analysis of information in Microblogs are Entity Mention Detection (EMD) and Entity Detection (ED). The state-of-the-art EMD systems aim to model the non-literary nature of microblog text by training upon offline static datasets. They extract a combination of surface-level features -- orthographic, lexical, and semantic -- from individual messages for noisy text modeling and entity extraction. But given the constantly evolving nature of microblog streams, detecting all entity mentions from such varying yet limited context of short messages remains a difficult problem. To this end, we propose a framework named EMD Globalizer, better suited for the execution of EMD learners on microblog streams. It deviates from the processing of isolated microblog messages by existing EMD systems, where learned knowledge from the immediate context of a message is used to suggest entities. Instead, it recognizes that messages within a microblog stream are topically related and often repeat entity mentions, thereby leaving the scope for EMD systems to go beyond the localized processing of individual messages. After an initial extraction of entity candidates by an EMD system, the proposed framework leverages occurrence mining to find additional candidate mentions that are missed during this first detection. Aggregating the local contextual representations of these mentions, a global embedding is drawn from the collective context of an entity candidate within a stream. The global embeddings are then utilized to separate entities within the candidates from false positives. All mentions of said entities from the stream are produced in the framework's final outputs. Our experiments show that EMD Globalizer can enhance the effectiveness of all existing EMD systems that we tested (on average by 25.61\%) with a small additional computational overhead.
\end{abstract}


% \fi
\IEEEpeerreviewmaketitle

\section{Introduction}
\label{sec:intro}
\begin{figure}[!t]
% \vspace{-0.35cm}
\centering
\includegraphics[height=1.9in, width=3.45in]{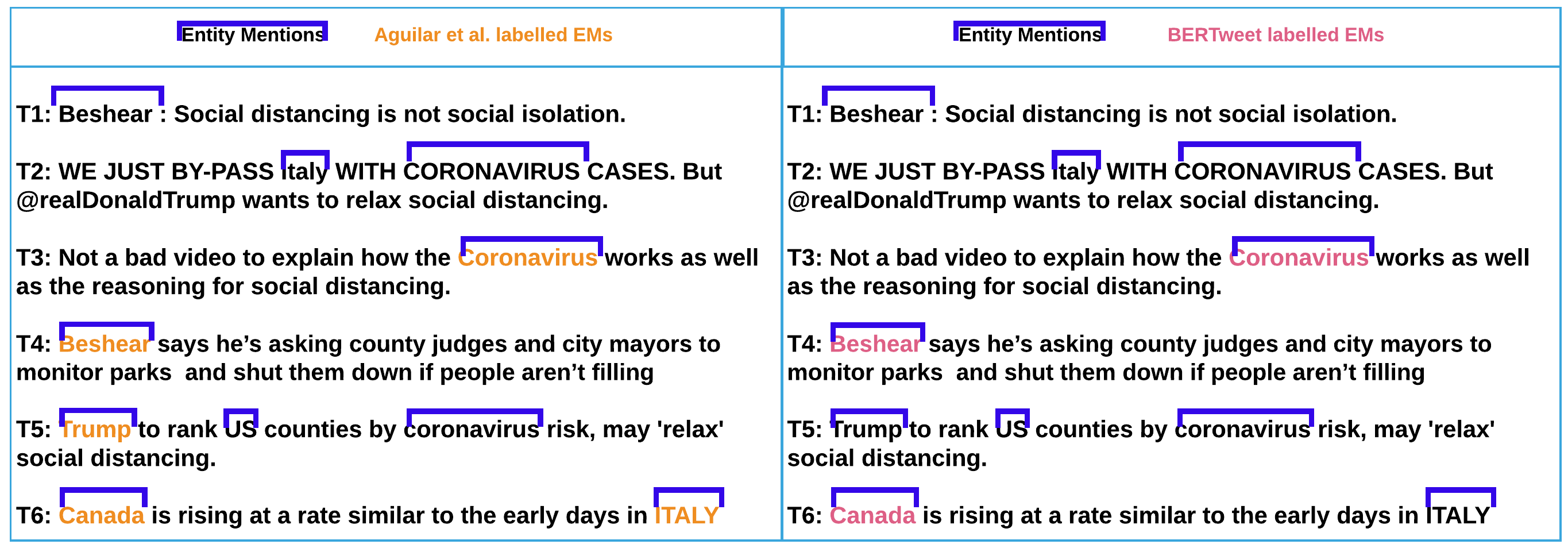}
%[height=2in, width=3.45in]
%[height=1in, width=2.5in]
%[scale=0.1]
\par
%\vspace{-8pt}
\caption{EMD on a message stream discussing Coronavirus}\label{fig:coronavirus-EMD}
\vspace{5pt}
\end{figure}

Entity Mention Detection (EMD) is the task of extracting contiguous strings within text that represent entities of interest. These strings (also known as surface forms) 
%(as described in WNUT17 \cite{wnut2017results}), 
are referred to as Entity Mentions (EMs). The benchmarking guidelines set by WNUT \cite{wnut2017results} identifies EMD and Entity Detection (ED) as two concomitant tasks in this context. 
%In the streaming environment,  
ED aims to cover the range of unique entities within text, while EMD compiles the %various surfaces forms or
string variations of entities from the text.
%EMD, along with entity typing (e.g., person, location, organization) are sub-tasks of the Named Entity Recognition (NER) problem. 
Together, they form the broader problem of  
%Together, with entity typing (e.g., person, location, organization), they constitute the
Named Entity Recogniton (NER) that has received significant research attention. In this paper, %we focus on EMD for microblog streams. In particular, 
we focus on maximizing effectiveness of state-of-the-art EMD techniques for the microblog streaming environment. 
% We first present some context for the problem.
\begin{myexample}
% \textcolor{blue}{
Tweets in Figure \ref{fig:coronavirus-EMD} have entity mentions (in many string variations) from six unique entities: `\textit{beshear}' in T1 and T4, `\textit{italy}' in T2 and T6, `\textit{coronavirus}' in T2, T3 and T5, `\textit{trump}' in T5, `\textit{US}' in T5 and `\textit{canada}' in T6.
% }
\end{myexample}
Off-the-shelf EMD solutions typically range from statistical Machine Learning models \cite{CenDSO13,Ritter,habib2015need4tweet,Finkel,SchneiderMD18} to Deep Neural Networks (DNNs) \cite{2017neuroner,aguilar-etal-WNUT17,ASU}.
%with the latter overwhelmingly populating the recent state-of-the-art.
However, the commonality among EMD systems focussing on Microblogs resides in their design and offline training process on static datasets. These datasets are curated from randomly sampled messages. As such, they provide a good representation of the non-literary language used in these platforms. Microblog EMD systems primarily study the nuances of its noisy text. They rightfully identify the lack of adherence to proper grammar, syntax or spellings as a key challenge to be addressed for language tasks. To extract contextual information from messages, they use a combination of surface level information -- word embeddings, lexical, orthographic and/or semantic information (POS tags, dependency trees), and sometimes even external gazetteers. 

For EMD from Microblog streams, existing systems do not take into account any of the streaming environment's defining traits and simply extend their processing approach for offline datasets. More precisely, these systems run each message in the stream through their EMD pipeline in isolation, one at a time, in the order of its entry into the stream. But given the underspecified context of a single, character-limited message, added to the constantly evolving nature of entity mentions within a microblog stream, detecting every mention of an entity from the stream remains a difficult problem to generalize. The varying textual context where EMs reside in messages is often further complicated by the rarity of many microblog-referenced entities in off-the-shelf lexical resources. This makes it more difficult to consistently extract mentions of novel entities for most EMD tools \cite{Ritter, Finkel}, including even the most effective Deep Learners \cite{aguilar-etal-WNUT17}. To illustrate these problems, we perform EMD on a message stream discussing the most prevalent conversation topic of 2020 -- the Coronavirus. We used two existing deep EMD baselines for this task: Aguilar et al. \cite{aguilar-etal-WNUT17} and BERTweet \cite{bertweet} (finetuned for EMD).

\textbf{A Case Study.} 
The objective of this study is to explore the performance of a state-of-the-art deep EMD tool on a microblog stream and understand its limitations.
%, developed for offline EMD, 
%in a streaming setting in order to understand how they can be mitigated. 
We run both baselines on a streaming dataset of 2K tweets ($D_2$, see Table 1 in Section \ref{sec:experiments}) generated from a Coronavirus tweet stream. We apply their production versions directly in this setting.
%, without any modification. 
BERTweet yields a modest performance on this stream subset with an F1 score of 53\%. 
% Aguilar et al., which uses more updated Twitter-trained word embeddings and gazetteers to overcome the limitations of deep EMD systems in tackling emerging entities, fared better at 60\%. 
% \textcolor{blue}{
Aguilar et al. fared better at 60\% with its reliance on updated Twitter-trained word embeddings and gazetteers, to better cover some rare entities. %Both systems exhibit 
%the same underlying issue. 
% The findings of this study are very revealing. 
%The moderate F1 scores are not only due to their 
Apart from missing some entities altogether, both systems show inconsistency in extracting EMs of the same entity throughout the stream, detected in some tweets but missed in others.%}

\textbf{Takeaways.} A closer look at the EMD results of both systems for tweets in the Coronavirus dataset in Figure \ref{fig:coronavirus-EMD} shows that they often missed mentions of one of the most important and frequent entities in this stream, i.e. `Coronavirus'. Other EMs that frequently came up in the stream but were also frequently missed include `Italy' and `US'. %Although Aguilar et al., which uses more updated Twitter-trained word embeddings and gazetteers to overcome the limitations of deep EMD systems in tackling emerging entities, fared better at 60\%.
%Despite Aguilar et al.'s usage of external resources, an entirely novel but frequently mentioned entity like `Coronavirus', absent in most vocabularies, still went undetected. %The failure to detect different variations of frequent entities, when they appear in different textual contexts within the stream, brings down its overall EMD effectiveness for the stream. 
For example, the entity `Coronavirus' has three mention variations in the tweets in Figure \ref{fig:coronavirus-EMD} but only `Coronavirus' was successfully detected while `CORONAVIRUS' and `coronavirus' were not. Note that the problem here is not merely the inability to identify the same entity string across different syntactic formats. It is rather the varying contexts (both syntactic and semantic) in which entities are repeated within a stream that adds to the challenge of generalizing an EMD algorithm that works well across all possible message contexts.

The appearance of entities in their many mention variations is an integral part of the microblog streaming ecosystem that constantly generates messages on multiple contemporaneous topics, i.e. conversation streams, evolving over time \cite{AljebreenMD21}. The failure to consistently identify these entity mentions leads to reduced EMD performance. 
%on the overall stream. Nonetheless, 
This appears to be a critical weakness of many state-of-the-art %deep 
EMD techniques. The common solutions to cope with this problem are more training or transfer learning (e.g., fine-tuning) on messages from newer topic streams. % but these are rather impractical. 
But as content in microblog streams evolves or spearheads into different topics, one needs to constantly annotate datasets specific to these emerging topics. This is not a scalable proposition. %To this end, we design a framework that can empower %deep 
%EMD systems to address the problem of inconsistent labeling in a more systematic manner. 
To this end, we design a framework that boosts the ability of EMD systems to recognize EMs more consistently across contexts.

\textbf{Local vs Global Context.} Majority of EMD systems encode the tokens of an input message with a variety of surface-level information to extract entity mentions. Pre-deep learning systems directly use this information to perform sequence labeling for EMD and identify entity boundaries within text using some variation of the BIO (beginning-inside-outside) encoding. 
Deep Neural Networks performing EMD however, use it to generate `\textit{contextual word embedding}' -- a representation of a token, in the context that it appears within a sentence. These contextual embeddings are then used for the downstream task of sequence labeling. % for EMD to identify entity boundaries using some variation of the BIO encoding.
Irrespective of architectural differences or the resources used to gather contextual information, these systems ultimately follow a traditional computation philosophy of treating individual microblog messages as isolated points of input that are individually processed to generate EMD outputs. We call this the `\textbf{Local EMD}' approach. This treats message streams as any other offline dataset -- an unchanging collection of messages, not a medium of incremental and topically-related message generation over an extended period of time. Given the various noisy but limited contextual possibilities of microblog messages, it can be untenable to individually analyze them and produce consistent labeling \cite{HeHMOD20}. This provides the motivation to move beyond the localized context of a message and
%follows this intuition to 
dynamically aggregate token-level local contextual embeddings from the entire stream, which are then used to derive a pooled `global' context for every token encountered within a dataset \cite{akbik-pooled-embeddings}. The global contextual embeddings are then concatenated to the sentence-level local embeddings for the eventual sequence labeling task. Expanding on this idea we argue that, more so than offline datasets, microblog streams are aptly positioned to collectively view embeddings. Messages within a conversation stream not only repeat a finite set of entities but also the context in which they appear, owing to inter-message topical correlation. Hence global contextual representations of tokens, or rather entity candidates, can be effectively mined and used for EMD in this setting.

\textbf{Approach Overview.} Here we propose the EMD Globalizer framework. It begins with a traditional EMD system that extracts local contextual information for each individual message and uses them to extract entities from messages. We call this step `\textbf{Local EMD}', due to the contextual information and inference drawn being locally confined. However, as evidenced before, local EMD tends to be inconsistent in providing the best message-level entity coverage. Hence its EM outputs are not instantly trusted. In our approach, the EM outputs from local EMD are used to generate a set of seed entity candidates. Additionally, in case of deep EMD systems, the token-level contextual word embeddings generated for every message are also stored. 
We follow this up by an occurrence mining step that finds additional mentions of seed candidates that are initially missed at Local EMD. Whenever an instance of a seed candidate is found, the local contextual information generated from its mention is aggregated to incrementally construct a global candidate embedding. 
For deep EMD systems, the contextual embeddings generated during Local EMD for the candidate's tokens are passed through a `\textit{Phrase Embedder}' that converts token-level embeddings into an embedding for the entire candidate string. 
Non-deep EMD systems, however, do not provide token-level representations and here we resort to extracting a syntactic embedding of the candidate mention depending on its immediate context as shown in \cite{twics}. %We can cite the TwiCS paper here.
Note that in either case, these candidate embeddings still capture only local contexts until this point. For all mentions of a seed candidate, the local candidate embeddings are aggregated to form a pooled global embedding. 
Global candidate embeddings are then passed through an `\textit{Entity Classifier}' to separate entities in the seed set from false positives that arose during Local EMD. All mentions of such discovered entities are considered valid entity mentions and produced as outputs. The steps following Local EMD up to the identification of true positives by the classifier together constitute what we call `\textbf{Global EMD}'. By decoupling the local EMD step from the global one, we arrive at a stream-aware EMD framework, %EMD Enhancer, 
that can plugin any existing EMD algorithm without training modification/fine-tuning and still enhance its EMD performance within a stream.

Our experiments show that EMD Globalizer effectively performs EMD, especially on microblog streams. We test it with four different EMD systems, including two state-of-the-art deep EMD networks, for local EMD. %: 1) Aguilar et al. \cite{aguilar-etal-WNUT17} -- a BiLSTM-CRF network, and 2) BERTweet \cite{bertweet} -- a BERT language model, pre-trained on tweets and then fine-tuned for EMD using the WNUT17 \cite{wnut2017results} training data. 
%In each case, the effectiveness of system was significantly boosted from it's standalone execution when plugged into the framework. The average improvement across all systems and datasets is 25.54\%
In each case, the effectiveness of an EMD system was significantly boosted (on average by 25.61\% across all datasets) 
when plugged into the framework. 
The framework also surpasses the best EMD performance on existing benchmarking datasets.
The uniqueness of the framework is that it can accommodate a variety of (local) EMD systems with no algorithmic modification and still achieve more consistent EM labeling across the stream. %When instantiated with sophisticated EMD taggers that prioritize model generalization capabilities, the framework can make their detection more robust to syntactic or semantic perturbations in Microblog text, thereby maximizing performance. However, it can also embolden non-complex techniques that prioritize near real-time EMD on Microblog streams without compromising on the eventual effectiveness.

This paper makes the following contributions:
\begin{itemize}[leftmargin=*]
\item We propose a novel framework for EMD in microblog streams. It consists of a Local EMD step, followed by a Global EMD step that includes an Entity Classifier. Our framework can accommodate both pre deep learning EMD systems and deep EMD systems and effectively collectivize EMD information in either case for Global EMD. It supports continuous and incremental computation which is in tune with the message generation process of streams.

\item The local EMD step is decoupled from the rest of the pipeline. This allows us to test the hypothesis of collectivising local embeddings to generate better performance for different (local) EMD systems. We demonstrate the framework's impact on several state-of-the-art instantiations of the Local EMD step. The contribution of this framework is therefore larger than catering to any single EMD technology.

\item We conduct extensive experiments using multiple Twitter datasets to test the proposed framework. We use both in-house streaming datasets to simulate EMD from Twitter streams and third party datasets curated from microblogs.
\end{itemize}

The complete implementation of EMD Globalizer and all the experimental datasets are available at \url{https://github.com/satadisha/collective-EMD-framework}.
\vspace{-5pt}
\section{Related Work}
\label{sec:RelWork}
The principal issues surrounding the EMD problem from Microblogs are identified in \cite{wnut2017results} to be the lack of annotated data from this domain, and congruently, the difficulty in identifying emergent entity forms. 
The EMD literature features a wide range of supervised systems
% The majority of EMD systems catering to this problem are supervised, 
with either handcrafted linguistic features  %\cite{Ritter,twitie,knn,drop-out-CRF,mishra2016-wnut-ner} 
in a non-neural system or DNNs %\cite{deepnnner-neural,liu-gazetteer-neural,yadav-deepNER-survey,2017neuroner,ASU,cambridgeLTL} 
%start with word embeddings and 
with minimal feature engineering. 
The first category of systems like \cite{Ritter,twitie,knn} recreate an information extraction pipeline in the form of POS-taggers or noun phrase chunker to extract and feed semantic features to a CRF based entity boundary segmentation module. In some systems \cite{cherry-unreasonable,hallym,toh,drop-out-CRF,iitp,tian-lattice,mishra2016-wnut-ner}, the feature set is further enhanced with word embeddings and gazetteers to better suit the limited yet diverse contextual possibilities of microblogs and the rare tokens that inhabit the medium. 
With advances made in Deep Learning, many deep neural networks (DNNs) \cite{transfer-learning-neural,deepnnner-neural,multimodal-neural,2017neuroner,shen2017deep-active, ZhangHDV19,ZhangHVD18} have been adopted for the sequence labeling task of NER. The recent WNUT shared task reports \cite{wnut16-report,wnut2017results} delve into a variety of neural architectures specifically designed for Entity Extraction from Tweets.
%Therefore, 
We have chosen Aguilar et al. \cite{aguilar-etal-WNUT17} -- a BiLSTM-CNN-CRF architecture that performed best at the WNUT17 task, and BERTweet \cite{bertweet} -- a BERT model trained on a large Twitter corpus that we finetune with the WNUT17 training data for EMD, as two of our Local EMD systems. %It has an implementation, available online for production use, to custom train and test new datasets.
% Some are however unsupervised \cite{Li:Twiner} or jointly model \cite{habib2015need4tweet, OpenCalais, gattani2013entity} EMD with other IE sub-problems.
\cite{domain-dependent-ner,domain-adap-ner} examine the cross-domain transferability of DNN features learned from the more abundantly labelled well-formatted corpora to overcome the lack of annotated data. %But handling of emergent entities largely remains to be addressed by enhancing architectural robustness of DNNs \cite{targetable-entities}.

Few other alternatives include TwiNER \cite{Li:Twiner}, an unsupervised approach using collocation features prepared from the Microsoft Web N-Gram corpus \cite{MicrosoftNGram} or joint modeling of NER \cite{NEN,yamada-EL,gattani2013entity} with another information extraction subtask. 

% \textcolor{blue}{
%A common theme for all the aforementioned systems however is to process input sentences individually and limit their feature extraction to the immediate local context.
%under consideration. 
%Up to now, 
The concept of globalizing EMD computation is encountered in
%has been limited to the setting of 
traditional NER pipelines for documents. 
Unlike a stream of tweets produced by multiple authors, documents are structurally more cohesive and contain well-formatted language. But both repeat entities and tokens through the collective span of their contents. Document-level EMD systems like HIRE-NER \cite{hire-ner}  %and DocL-NER \cite{docL-ner}
utilize this tendency to distill non-local information for each unique token, from the entire scope of the document, using a memory-structure and append them to sentence-level contextual embeddings before an EMD decoder draws final output labels. 
DocL-NER \cite{docL-ner} additionally includes a label refinement network to enforce label consistency across documents and improve EMD results. 
We compare EMD Globalizer with HIRE-NER \cite{hire-ner} to test how effectively global information for EMD is compiled in each system.%}

% EMD Globalizer distinguishes from these systems in several ways. 
% \textcolor{blue}{
For microblogs, TwiCS \cite{twics} explores the feature  
of entity mention repetition 
to efficiently perform EMD on 
%voluminous 
streams. Using a shallow syntactic heuristic it identifies entity candidates and collectively generates syntactic support from their mentions within the entire stream to distinguish legitimate entities from false positives.%}

% \textcolor{blue}{
Our proposed EMD Globalizer further extends this idea of looking beyond the modeling of singular sentences in a stream. 
It combines the potential of collective processing of microblogs from a stream %introduced in TwiCS 
with existing EMD techniques that, despite offering robust EMD generalization, are constrained to processing sentences individually. 
Unlike TwiCS and other standalone EMD systems, 
what we propose in this paper is not a standalone system but a general EMD framework that aims to enhance the effectiveness of existing (local) EMD systems, 
%that we call `Local EMD', 
when applied to microblog streams.%}
%Extending on an initial Local EMD system, the framework uses incremental occurrence mining to collate the various mention-level contexts of an entity %from all sentences. % and analyze them collectively. 
%This ensures the collective utilization of global information 
%from the larger scope of the entire stream in a step we call `Global EMD'.  The collective context is called a `global entity embedding' that can be used to generate final EM outputs. 
%The resulting EMD process is not constrained to a particular type set and considers entity types (e.g., People, Location and Organization) predominantly covered in most EMD systems. 
% \textcolor{blue}{
Furthermore, in this work, we integrate a diverse set of Local EMD systems into our framework, with a special design focus on Deep EMD learners. TwiCS \cite{twics} is not featured as a Local EMD system in our experiments since it does not process sentences individually.%}
\begin{figure}[!t]
%\vspace{-2pt}
\centering
\includegraphics[height=2.225in, width=3.25in]{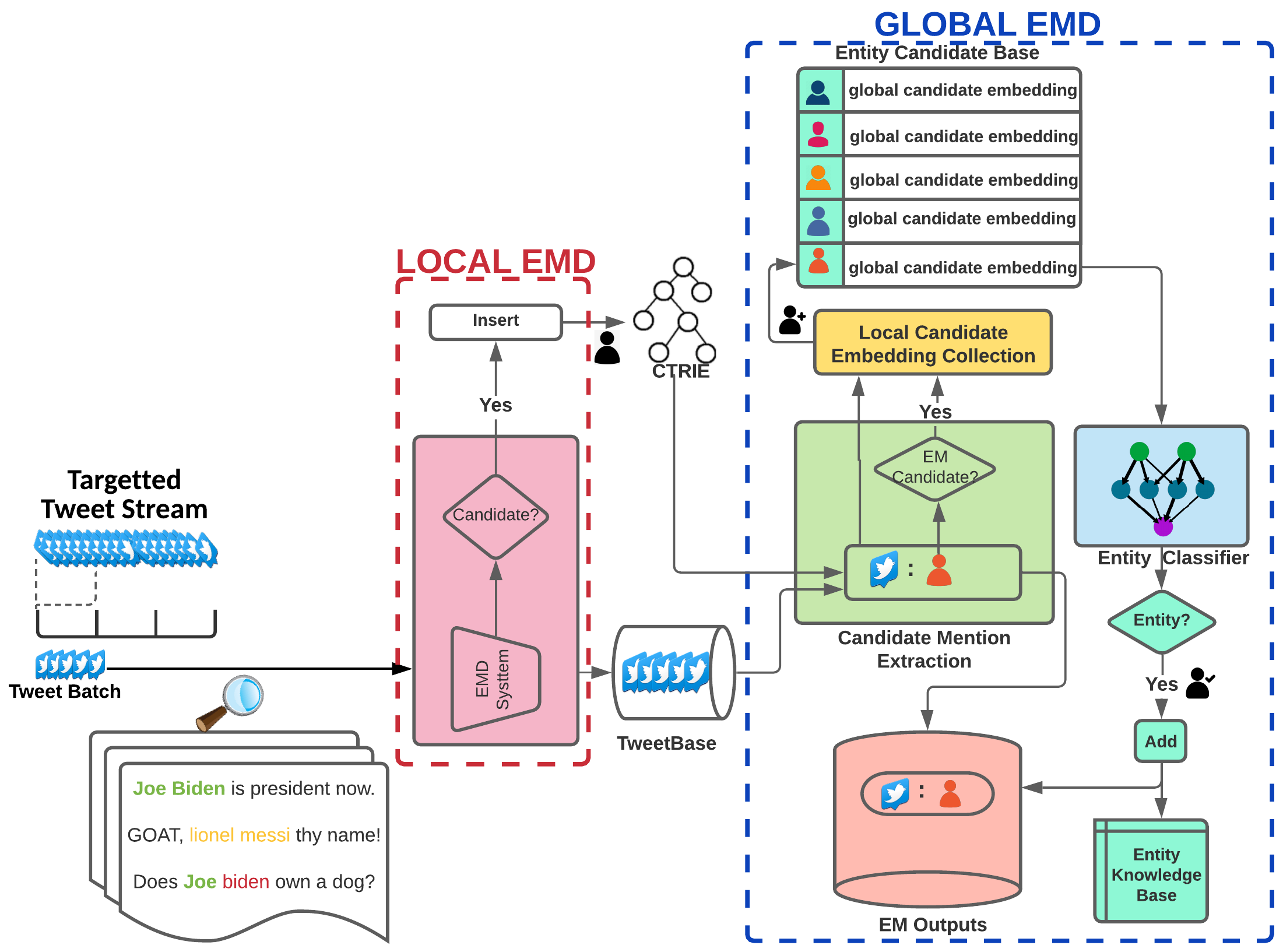}
%[height=2.35in, width=3.35in]
%{syem-execution-sequence-New1.png}
%[height=2.5in, width=2.5in]
%\vspace{-15pt}
\caption{EMD Globalizer System Architecture}\label{collective-emd-execution}
\vspace{2pt}
\end{figure}

\section{System Architecture}
\label{sec:system-overview}

Figure \ref{collective-emd-execution} illustrates the overall architecture of EMD Globalizer. Note that, depending on the type of system employed for Local EMD, the components of the rest of the framework are adjusted accordingly. %This framework facilitates continuous execution of a tweet stream over multiple iterations (or batches).  A batch is simply a set of tweets in the same stream that will be processed together. 
The EMD process in this framework facilitates continuous execution of a tweet stream over multiple iterations. Each iteration consists of a batch of incoming tweets thereby discretizing the evolution of messages within the stream. A single execution cycle through this framework can be divided into the following steps:\\
\textbf{(1)} The Streaming module fetches a stream of tweets, on a particular topic, using the Twitter streaming API.\\
\textbf{(2)} First we run a batch of tweets through an off-the-shelf (deep or non-deep) EMD system, one sentence at a time, in the \textbf{Local EMD} step. Phrases labelled as possible entities are registered as entity candidates. Further, in case of a deep EMD system, the output of the neural network's final layer before token-level classification are stored, for every tweet in the batch, as `\textit{entity-aware embeddings}' of sentence tokens.\\
%In addition, for every tweet in the batch, the output of the deep neural network's final layer are stored, prior to token-level classification with BIO output labels, as `entity-aware embeddings' of tokens in a tweet.\\
\textbf{(3)} Next we initiate \textbf{Global EMD}. This includes a few added steps where individual framework components are adjusted according to the type of local EMD system inserted into the framework: (i) First, an additional scan of the tweet batch extracts all mentions of the entity candidates that have been discovered so far. This involves finding candidate mentions that were missed by the local EMD system in the previous step, along with the ones that were already found. (ii) For every mention we find, 
a candidate embedding is constructed based only on the immediate local information. With a regular (i.e., non-deep) local EMD system, we construct a syntactic embedding for the candidate mention from its immediate context. With a deep local EMD system however, 
token-level contextual embeddings are also available at the end of the local EMD. Hence, in this case, the token-level embeddings for the candidate mention phrase are passed through the Phrase Embedder to obtain a unified contextual embedding for the entire phrase. %\\
(iii) Local candidate embeddings of every mention of an entity candidate found within a batch of tweets are aggregated to generate the candidate's pooled global embedding. The global embedding can be incrementally updated by adding local embeddings into the pool as and when new mentions arrive. %\\
% \textbf{(4)} 
(iv) The final step is to pass global candidate embeddings through the entity classifier to separate the entities within the seed candidates from non-entities. Mentions of candidates that get labelled as entities are produced as valid entity mentions in the system's final EMD outputs for the tweet batch.

\begin{figure}
%\vspace{-2pt}
\centering
\includegraphics[height=2.225in, width=3.25in]{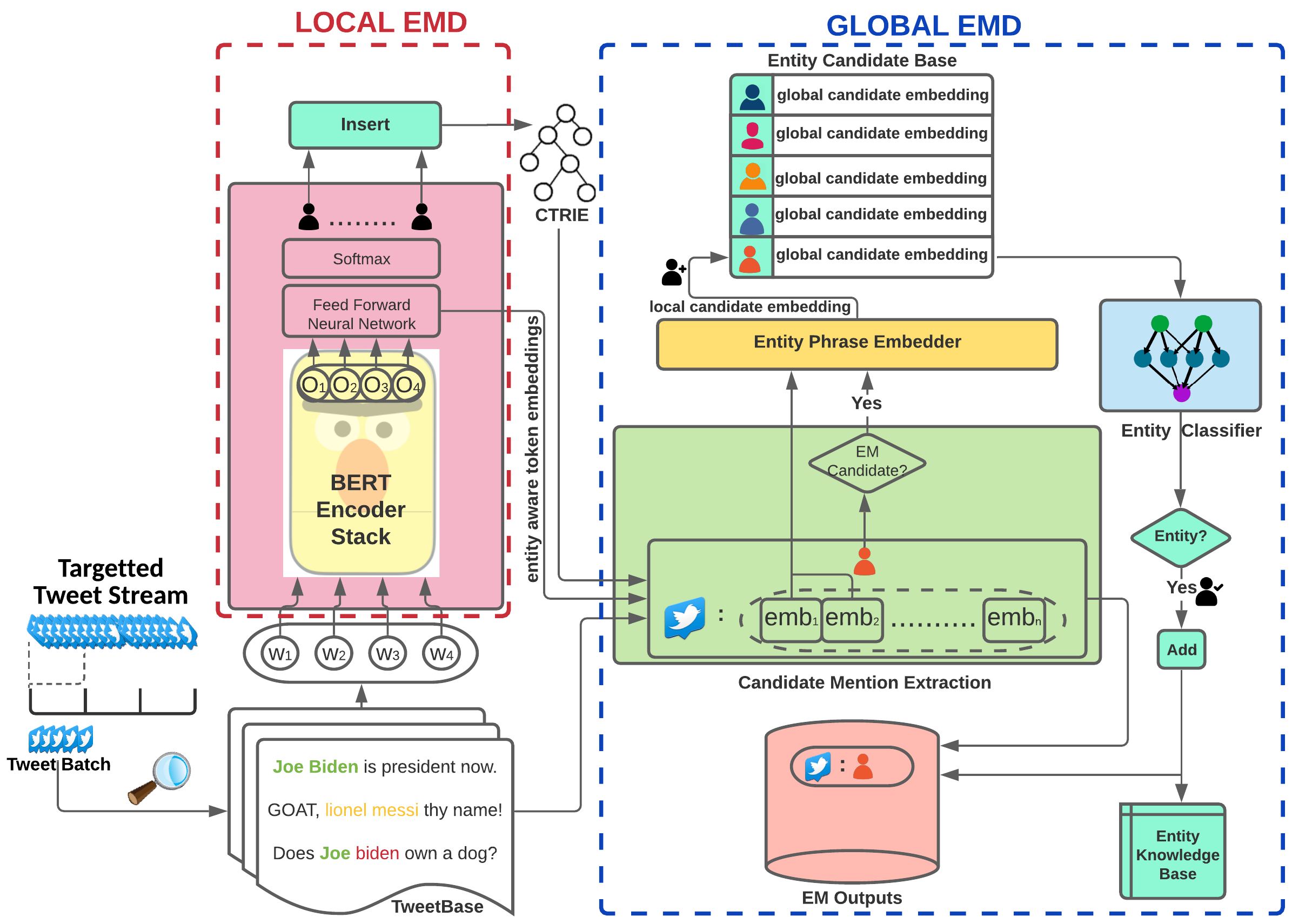}
%[height=2.35in, width=3.35in]
%{syem-execution-sequence-New1.png}
%[height=2.5in, width=2.5in]
%\vspace{-15pt}
\caption{EMD Globalizer with a Deep EMD System}\label{collective-emd-execution-deep}
\vspace{1pt}
\end{figure}

We elaborate on these steps in later sections. The framework, when initialized with a deep neural network for local EMD, consists of a few additional steps. We zoom into it separately from the overall system architecture in Figure \ref{collective-emd-execution-deep}.
\section{Local EMD}
\label{sec:local-emd}
The Local EMD step can be any existing EMD algorithm that processes every single tweet-sentence in a stream, or in a tweet batch, individually and indicates likely entity mentions. 
A variety of existing systems can be plugged into the framework at the local phase. 
For majority of these systems, the EMD process is designed as a sequence labeling task where each token is tagged relative to its nearest entity boundary by adopting a variation of the BIO encoding. To facilitate a token's labeling, local information relative to the token is generated and used. In case of deep learners, this happens to be a token-level contextual embedding obtained at the penultimate layer of the deep neural network, prior to the generation of output labels. 
For non-deep systems, this can be rich token information like POS-tags or gazetteer features that can aid the labeling process. 
Therefore, we interpret it as an encoding of the local entity-aware information, extracted from within the context of a single input.  %The purpose of local EMD is therefore to both suggest entity candidate and to encode local, token-level entity information for every tweet in the batch.

\smallskip
\noindent \textbf{Objectives}: For a targeted stream of tweets, Local EMD aims to: (1) identify substrings from individual sentences as mentions of potential entities, (2) encode local entity information for every token in individual sentences, and (3) generate a set of seed entity candidates from tagged mentions.

\subsection{Instantiations}
\label{local-EMD-instantiations}
There are different ways to instantiate the Local EMD step. For EMD Globalizer, most off-the-shelf EMD systems that process sentences individually would qualify. We test with four EMD systems, each of which supports a different EMD extraction algorithm, including two state-of-the-art deep learners. We now briefly describe each of these instantiations. Note that Local EMD systems are inserted as blackbox within the framework without any technical alteration during testing.

\smallskip
\noindent
{\bf 1. Chunker Based EMD -- TweeboParser:~}
% \subsubsection{\textbf{Chunker Based Tagging: TweeboParser}} 
This Local EMD system is a dependency parser \cite{tweebo} trained on English Tweets drawn from the POS-tagged tweet corpus of \cite{owoputiCorpus}. We use the production version of TweeboParser to generate POS-tags and dependency trees that capture the semantic structure of tweets. We implement a procedure (NP Chunker) that extracts noun phrases from the parser-generated dependency trees and pass them as entity candidates to the rest of the framework.

\smallskip
\noindent
{\bf 2. CRF Based Tagging -- TwitterNLP:~} 
TwitterNLP \cite{Ritter} recreates the Information Extraction pipeline for tweets by rebuilding classical NLP tools with microblog-specific considerations. The system begins by building modules for 2 subtasks, that help draw shallow semantic features from the tweet text -- 1) Part-of-Speech tagging and 2) Noun Phrase Chunking. These features are later re-used for the actual NER task. The POS tagger (T-POS) and shallow parser (T-CHUNK) are CRF-trained models with additional tweet-sensitive features like @tags, hashtags, and URLs.  In addition to shallow parsing, TwitterNLP also constructs a capitalization classifier (T-CAP) to deal with the unreliable capitalization in tweets. The classifier is trained as an SVM and it studies capitalization throughout the entire sentence to predict whether or not it is informative. For EMD, TwitterNLP splits the subtasks of segmenting and classifying Named Entities in two separate modules. T-SEG, a CRF-trained discriminative model, that adopts BIO encoded labels, for entity level segmentation of tweets. The feature set for this task uses orthographic features like capitalization, contextual features gathered from T-POS and T-CHUNK, output of T-CAP as well as dictionary features, including a set of type-lists gathered from Freebase \cite{Freebase}, an open-domain ontology, and Brown Clusters. T-SEG uses a manually annotated training set for supervised learning of model parameters. We work with the production version of TwitterNLP available in Github for our experiments.

\smallskip
\noindent
{\bf 3. Multi-task Deep Neural Network -- Aguilar et al. \cite{aguilar-etal-WNUT17}:~} The best-performing system for the WNUT 2017 NER task \cite{wnut2017results} is an effective deep EMD system. It is primarily a BiLSTM-CNN-CRF network that follows the philosophy of multi-task learning. It is multi-task in the sense that it learns higher-order feature representations along three different tasks, each of which includes relevant information for the subsequent token-level entity tag prediction:\\
a) Character-level representation: character encodings from \cite{ma-bilstm-cnn-crf} and an orthographic encoder are fed to a Convolutional Neural Network to learn character-level representations. \\
b) Token-level representation: here both word level representations and POS representations are concatenated in a unified vector to denote token-level features. Word embeddings from \cite{godin-embeddings} are fed to a BiLSTM to learn word level representations, while a POS encoder is trained using POS tags obtained by parsing the text using TweeboParser \cite{tweebo}.\\
c) Lexical representation: tokens appearing as entities in select gazetteers \cite{mishra2016-wnut-ner} are encoded into a 6-dimensional vector -- one dimension for each gazetteer type. These lexical vectors are fed into a fully-connected layer with ReLU activation function.

The concatenation of these feature vectors is then fed to a common dense layer with a sigmoid activation function. Finally a CRF layer learns dependencies between the neural output nodes and conducts sequential labeling. 
The token level encoding scheme used for sequence labeling is BIO. We use the production version of the system available online.

\smallskip
\noindent
{\bf 4. BERTweet for EMD -- Nguyen et al. \cite{bertweet}:~} Pre-trained language models have become the go-to choice for many NLP tasks and several recent systems have adopted a pre-trained BERT language model that is fine-tuned for the downstream sequence labeling task of EMD. For our Local EMD instantiation we use BERTweet, the first large-scale language model trained on English Tweets. This system has the same architecture as BERT\textsubscript{base} \cite{bert-main-paper} but uses the RoBERTa \cite{roberta} pre-training procedure for more robust performance. The pre-training dataset is a collection of 850M tweets, each consisting of at least 10 and at most 64 tokens. fastBPE is applied to segment all tweets into subword units, using a vocabulary of 64K subword types. On average two subword tokens are maintained per Tweet. 
To fine-tune the language model for EMD, a feed forward neural network (FFNN) layer and a softmax prediction layer are added on top of the output of the last Transformer encoder. The fine-tuning is independent of the underlying BERT model's training. It is repeated five times with randomly initialized seeds. The reported performance is the average of five test runs. 
The BERTweet repository is publicly available on GitHub. We use a pre-trained BERTweet model available at the Hugging Face model hub that amasses a large collection of pre-trained language models catering to a variety of downstream NLP tasks. 

Every Local EMD system suggests a set of seed entity candidates derived from the EMs that are tagged and discovered by it. These seed entity candidates are stored in a \textbf{CandidatePrefixTrie} (\textbf{CTrie} for short). CTrie is a prefix Trie forest for efficient indexing of candidates. It also facilitates subsequent lookups for finding all mentions of discovered entity candidates later during the Global EMD phase. CTrie functions like a regular Trie forest with individual nodes corresponding to a token in a candidate entity string. %A path from the root to a leaf forms a full entity candidate string. 
Entity candidates with overlapping prefixes are part of the same sub-tree in the forest. Another data structure produced at the end of Local EMD is \textbf{TweetBase}. It maintains an individual record for every tweet sentence indexed by a (tweet ID, sentence ID) pair and a list of detected mentions that get updated as the sentences pass through Global EMD. 

In addition, deep EMD systems provide token-level contextual embeddings for each tweet-sentence in the input stream. These are also recorded for the computation of local candidate-level embeddings that are then used to generate global candidate embeddings. The token-level embeddings are collected from the final, pre-classification layer of deep EMD. For Aguilar et al. this would be the output of the last fully connected layer, prior to the CRF layer. For BERTweet, this would be the layer prior to the output softmax layer. 
In either case, these embeddings encode information that demarcates entity boundaries within the sentence tokens. Hence we call them local `\textit{entity-aware}' embeddings.

\section{Global EMD}
\label{sec:global-emd}
At the end of Local EMD an initial entity extraction is accomplished for every tweet-sentence in the TweetBase, sometimes with the generation of token embeddings that are aware of adjacent entity boundary information. The Local EMD outputs suggest a set of seed entity candidates stored in the CTrie. However Local EMD is prone to inconsistent detection of the same entity across the breadth of a stream. Mentions of entity candidates are missed in some sentences while detected in others. Here we introduce the Global EMD module to address some of these inherent limitations. More specifically, the purpose of Global EMD is to shift the focus of its EMD computation, beyond the confines of a single sentence. It views mentions of a candidate collectively, across the entire span of a stream, before determining if it is an entity.

\smallskip
\noindent \textbf{Objectives}: 
The Global EMD step addresses:\\
\indent\textbf{1. Removal of False Negatives}: False Negatives happen when Local EMD fails to tag 
true EMs. 
%were not tagged by Local EMD. 
For example, in Figure \ref{fig:coronavirus-EMD}, \textit{coronavirus} in T4 is a false negative.

\indent\textbf{2. Removal of False Positives}: False Positives happen when Local EMD extracts non-entity phrases 
%are extracted 
as entity candidates.
%by Local EMD.

\indent\textbf{3. Correction of Partial Extraction}: Partial extractions happen usually due to mislabeling of multi-token entities, where a part of an entity string is left outside of an entity window under consideration. %\textit{Carter page} in T4 in Figure \ref{fig:local-emd}.
Correcting such partial extractions improves both recall and precision.

\smallskip
\noindent \textbf{Execution}: 
The execution with Global EMD is broken down into three separate components. First, an additional scan of the tweet-sentences alongside a lookup through the CTrie, reveals all existing mentions of entity candidates, including ones that were previously missed by Local EMD. For every candidate mention encountered here, a local candidate embedding is extracted and recorded to its entry in a data structure called the `CandidateBase'. 
Depending on the local EMD system, the process of collecting local embedding varies. 
Non-deep systems do not generate token-level contextual embeddings along with their EM suggestions for a sentence. In this case, we generate a syntactic embedding for the candidate mention found from its immediate local context in a sentence.
For deep EMD systems, token embeddings are collected from the TweetBase entry of a sentence recorded at the end of Local EMD. Next, a candidate's token embeddings are together fed to an Entity Phrase Embedder to generate a unified local contextual embedding for the entire phrase, for this mention of the candidate. 
% For every candidate, mention-level local embeddings are collected and added to its entry in a data structure called the `CandidateBase'. 
Then, a pooling operation on mention level contextual embeddings gives us the `global candidate embedding'. Note that we update the global embedding of a candidate incrementally as we find new mentions in the stream. 
Finally, an Entity Classifier receives the `global candidate embeddings' to label every candidate as an `entity' or a `non-entity'. Candidates recognized as entity find their mentions produced as valid EMs in the final output for the stream. We now describe the components of Global EMD in more detail.

\subsection{Candidate Mention Extraction}
\label{subsec:candidate-mention-extraction}
In theory the purpose of the Candidate Mention Extraction step is to detect EM boundaries within text. Most EMD systems \cite{Finkel,knn} treat this as a sequence labeling problem. However, empowered by the seed candidates from Local EMD registered in the CTrie, the problem of segmenting tweets into candidate (EM) boundaries here is simplified to that of a lookup in the CTrie. The module analyzes every token in a tweet sentence, in conjunction with a CTrie traversal. With a case-insensitive comparison of tokens with CTrie nodes, this results in two possibilities:\\
%The subsequent problem can be formulated based on the following traversal scenarios of CTrie, posed by the kind of token Rescanning algorithm finds:\\
%\begin{enumerate}
% \indent\textbf{(i)} A token whose first letter is capitalized.\\
\indent\textbf{(i)} A token that matches a candidate node on the current CTrie path, when cases are ignored.\\
\indent\textbf{(ii)} A token matching no node in current path.
%\end{enumerate}

\textbf{The problem} is to check if a token forms a candidate mention alone or together with up to $k$ tokens following it. %Note that the case of partial capitalization can appear both in the prefix or suffix of a true positive candidate. %The outcome of the process is shown in Figure 

%As shown in Figure \ref{fig:rescan}, candidate mentions extracted during Phase I are verified, and sometimes corrected, via a syntax agnostic Rescan module.
The extraction process scans a tweet-sentence and identifies the set of longest possible subsequences matching with candidates in the CTrie, while case is ignored (e.g., "coronavirus" is a match for "Coronavirus"). As a consequence, candidate mentions extracted during Local EMD are verified, and sometimes corrected. For example, if the Local EMD system finds only a partial excerpt `\textit{Andy}' of the EM `\textit{Andy Beshear}' in a tweet, but nonetheless recognized the entire string in other tweets, the candidate (`\textit{andy beshear}') will be registered in the CTrie. This partial extraction can now be rectified and corrected to the complete mention. %Algorithm RescanMentions (alg. \ref{rescan}) compiles the various considerations of this module. 
The resulting process is syntax agnostic. It initiates a window that incrementally scans through a sequence of consecutive tokens. In each step it checks:\\
\indent\textbf{a)} whether the subsequence within the current scan window corresponds to an existing path in the CTrie. %(lines 13-24) 
If true, it implies that the search can continue along the same path, by including the token to the right within the window in the next iteration.\\
\indent\textbf{b)} whether the node on this path, that corresponds to the last token of the subsequence, refers to a valid candidate. %(lines 15-18) 
%-- this is a sub-condition of (a). 
If true, it implies that the subsequence can be recorded as the current longest match, before next iteration begins.

%Note that the subsequence match is syntax agnostic. This renders the aforementioned scenarios (i) and (ii) to be equivalent in terms of the scan. 
In case of a mismatch, i.e. scenario (ii), the module stores the last matched subsequence within the current window, and then skips ahead, by initializing a new window from the position of the token, next to it. %(lines 26-31). 
The search for a new matching subsequence path is initiated from the root of the CTrie. However, if the last search had failed to yield a match with any of the existing candidates in CTrie, the new window is initialized from a position that is to the immediate right of the first token in the previous window. %(lines 32-35). 
The process is repeated until all tokens are consumed. In the end we obtain a collection of mention variations for each entity candidate.

\subsection{Local Candidate Embedding Collection}
\label{subsec:local-candidate-embedding}

The process of collecting candidate embeddings from the local context of a candidate's mention depends on the type of Local EMD system. Here we broadly categorize them into two groups: 1) pre-deep learning systems-- systems that do not use (deep) neural networks
%i.e. non-deep 
EMD systems 
%that use statistical machine learning 
and 2) deep EMD systems-- systems that use deep neural networks. For the former, EMD execution is limited to generating BIO labels of sentence tokens that suggest entities. Hence, for these systems, we provide a workaround to study the immediate context of mentions and generate a local candidate embedding. Deep EMD systems, however, provide token-level contextual embeddings that encode latent relations between tokens in a sentence. Hence for deep learners we fully utilize them when preparing candidate embeddings.

\subsubsection{\textbf{Syntactic Distribution for Non-deep Local EMD}}
\label{subsec:syntactic-distribution}

For non-deep systems %we only get candidate entities at the end of Local EMD. To circumvent the lack of further information, 
we follow \cite{twics} %put in TwiCS's citation here
and extract an embedding that encodes the local syntactic context of a candidate mention into an embedding of 6 dimensions.
It enumerates all the syntactic possibilities in which a candidate can be presented.\\
% It enumerates the local syntactic contexts of mention variations of an entity candidate into six features.\\
(1) \textbf{Proper Capitalization}: This corresponds to the first character of every candidate token capitalized.\\
%Capitalization instances that the CS heuristic is capable of identifying. 
%This included first letter capitalization format for candidates appearing in a tweet sentence that is neither fully capitalized nor non-capitalized, with unigram candidates appearing not appearing at the start of sentence.
(2) \textbf{Start-of-sentence capitalization}: A unigram candidate capitalized at the start of sentence.\\
(3) \textbf{Substring capitalization}: Only a proper substring of a multi-gram candidate is capitalized.\\
(4) \textbf{Full capitalization:} Abbreviations like \textit{`UN'} or \textit{`UK'} where the entire string is capitalized.\\
(5) \textbf{No capitalization}: The entire string is in lower case.\\
(6) \textbf{Non-discriminative}: A sentence is entirely in upper or lowercase, or has first character of every word capitalized, so candidate mentions found are not syntactically informative.

In the end, the local syntactic embeddings are aggregated and pooled to derive a global candidate embedding. %What we obtain is a syntactic distribution of an entity candidate over all the six syntactic possibilities in which it can produce mentions.

\subsubsection{\textbf{Entity Phrase Embedder for Deep Local EMD}}
\label{subsec:entity-phrase-embedder}

The `entity aware embeddings' generated by a Local Deep EMD system are for individual tokens. However, the Entity Classifier verifies candidates based on their global contextual representation generated by aggregating local contextual representations of their mentions. 
So we need semantically meaningful representations of candidate mentions before they can be aggregated. Given that entity candidates have variable number of tokens, we need to combine the token-level embeddings into a unified, fixed-size embedding of the candidate phrase. This is the role of the Entity Phrase Embedder.

%While the NLP literature does not explicitly cover phrase embeddings, a similar well-researched problem is the generation of sentence embeddings for Semantic Textual Similarity (STS) tasks. 
To generate phrase embeddings, we refer to the work on sentence embedding for Semantic Textual Similarity (STS) tasks. Since the contextual embeddings provided by language models are token-level, the intuitive solution for a sentence embedding sourced from multiple token embeddings is to add an average (or max) pooling operation and arrive at an average embedding to represent the sentence. Alternatively, one can add a CLS (classification) token at the end of sentences and train them for a Natural Language Inference task. The embedding for the CLS token would be considered representative of the entire sentence. We however follow the approach in  %state-of-the-art system for sentence embedding generation, 
Sentence-BERT or SBERT \cite{sentence-BERT}. SBERT argues that using the aforementioned approaches do not yield good sentence embeddings, and can often be worse than averaging Glove embeddings \cite{glove-embeddings} of tokens.

SBERT uses a `\textit{siamese network structure}' and trains it for different STS tasks, including sentence similarity. The input set in this case consists of pairs of sentences whose similarity is calculated such that sentences conveying similar content have a higher score than those that do not exhibit any content similarity. A siamese network consists of two identical sub-networks that have the same architecture and parameters to generate representations of pair-wise inputs that are then compared to generate a similarity score which is the network's output. %The loss calculated on the output is back-propagated and the parameter updation is mirrored across both sub-networks. 
SBERT, in its implementation, uses the same BERT model as sub-networks in its siamese structure. It also adds an average pooling layer that generates a mean sentence-level embedding from the token-level embeddings of the BERT. Finally, the Cosine Similarity function is used to %compare the embeddings of pairwise sentence inputs by 
generate a similarity score upon which the loss function is calculated. Mean squared error loss is used as the regression objective function to train SBERT for this task. The updation of weights during back-propagation is mirrored across both sub-networks.

\begin{figure}
%\vspace{-2pt}
\centering
\includegraphics[height=1.95in, width=2.05in]{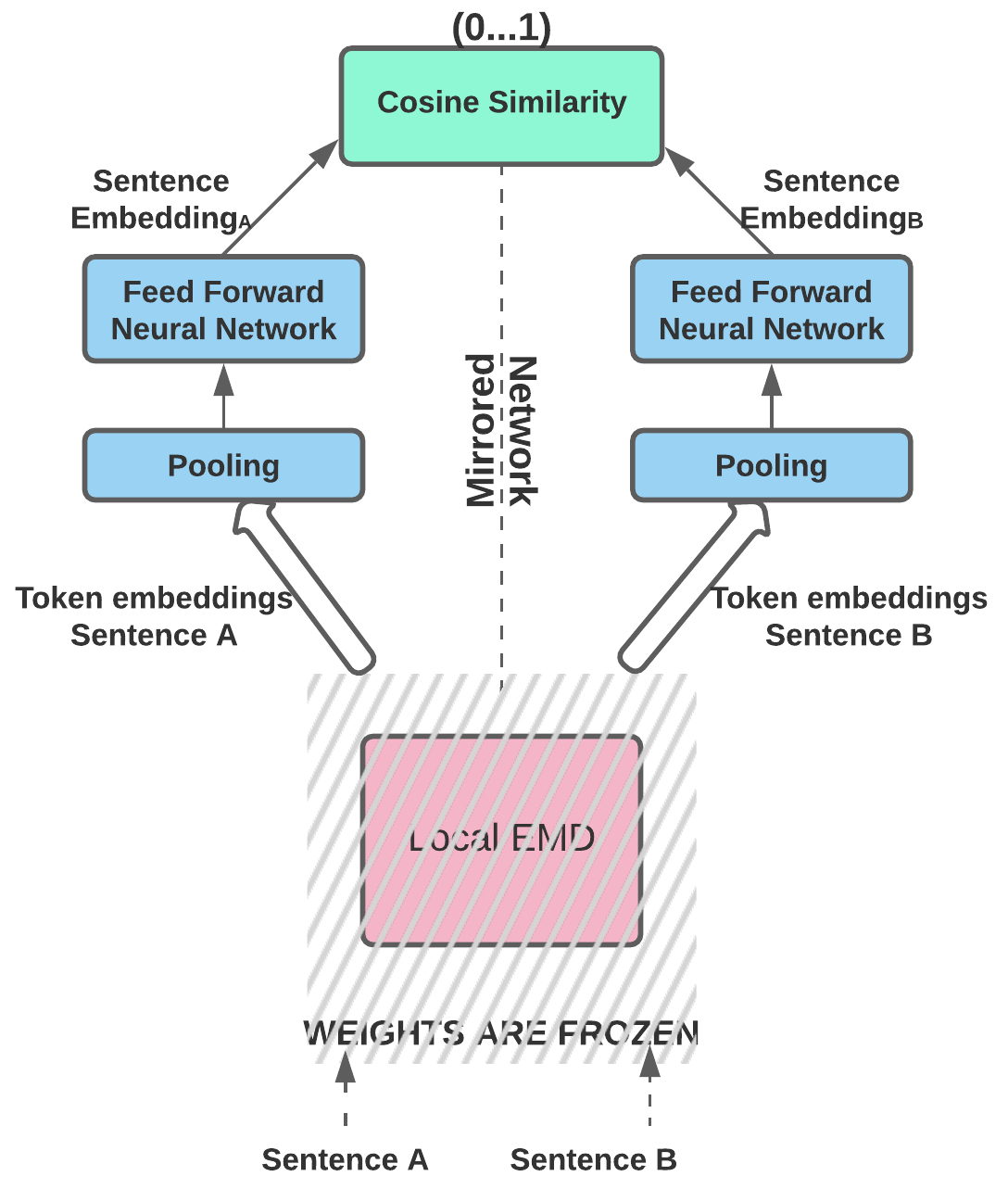}
%[height=2.35in, width=3.35in]
%{syem-execution-sequence-New1.png}
%[height=2.5in, width=2.5in]
%\vspace{-15pt}
\caption{Entity Phrase Embedder in Modified Siamese Network}\label{fig:entity-phrase-embedder}
\vspace{1pt}
\end{figure}
The Entity Phrase Embedder used in our framework is shown in Figure \ref{fig:entity-phrase-embedder}. It follows a modified design of SBERT and is also trained on the sentence similarity task. We use the deep neural network used for Local EMD to generate token-level embeddings as the principal component of the mirrored subnetwork in our siamese structure. In addition we add an average pooling layer to combine token-level representations into an average embedding that is then passed on to a dense layer. The Cosine similarity score between the dense layer outputs of the two subnetworks gives the final output upon which the regression loss function is calculated and backpropagated. Unlike SBERT however the gradient computation is not backpropagated all the way back to the deep neural network (the BERT engine in case of SBERT). In other words, the DNN's weights remain frozen in our siamese network %and parameter updation is limited to the layers added on top of it. 
and only the weights of the layers following it are updated. 
This is because the deep neural network's role in our framework is to produce (Local) EMD results for which it had already been optimized. The rest of the subnetwork however work on the 'entity-aware' token embeddings to produce an optimal sentence/phrase embedding that performs well on the semantic similarity task. The dense layer is also useful to customize the phrase embedding size, separate from the underlying token embedding size generated by the DNN. For example, when training the Entity Phrase Embedder with BERTweet, where token embeddings are of 768 dimensions, we bring down the candidate embeddings from the Entity Phrase Embedder to 300 dimensions. 
%Like SBERT, we train the Entity Phrase Embedder using the STS Benchmark (STS-b) data \cite{sts-benchmark}. %An illustration of this process using a modified siamese network is given in Figure \ref{fig:entity-phrase-embedder}. 
Further details on the training process of the Entity Phrase Embedder is provided in Section \ref{sec:training-phrase-embedder}. %We use Adam optimizer \cite{adam} with a fixed learning rate of 0.001 and batch size of 32. %We also enable early stopping in our training. 

% \textcolor{blue}{
A candidate's local embedding ($local\_emb \in \mathbb{R}^{d}$) from its token-level embeddings can be computed using one of the Entity Phrase Embedder sub-networks as
\begin{equation} \scriptsize
    pooled\_emb = \frac{1}{\left|T \right|}\sum_{j=1}^{\left|T \right|}token\_emb_{T_{j}}
\end{equation}
\begin{equation} \scriptsize
    local\_emb=W_{ff}pooled\_emb+b_{ff}
\end{equation}
\normalsize
where $T$ denotes the set of tokens in the candidate phrase, $token\_emb_{T_{j}} \in \mathbb{R}^{d}$ is the contextual embedding of the j-th token in $T$. The weight matrix $W_{ff}\in \mathbb{R}^{d \times d}$ and bias $b_{ff}\in \mathbb{R}^{d}$ are trainable parameters from the mirrored sub-networks. %}

\begin{figure}[!t]
%\vspace{-2pt}
\centering
\includegraphics[height=1.8in, width=3.45in]{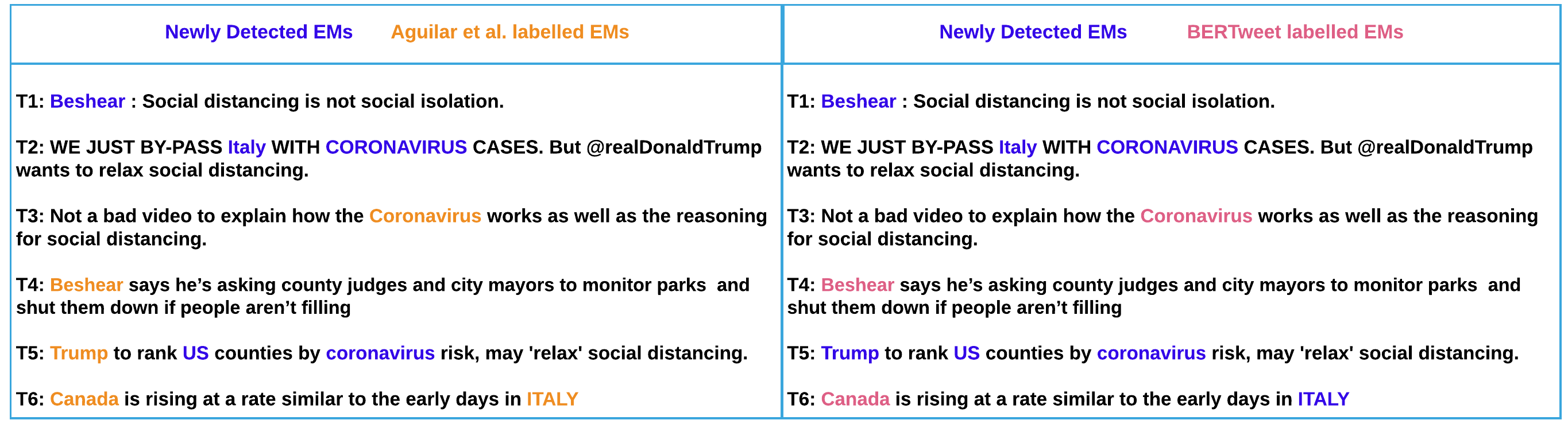}
%[height=1.8in, width=3.45in]
%[scale=0.14]
\par\vspace{1.5pt}
\caption{EMD Globalizer on a message stream on Coronavirus}\label{fig:coronavirus-EMD-followup}
\vspace{2pt}
\end{figure}

\subsection{Entity Classifier}
\label{subsec:entity-classifier}
The information encoded in the local embeddings of individual mentions of a candidate is limited to the context of the sentence containing it.
We add these local embeddings to the candidate's record in a data structure called the CandidateBase, which maintains an entry for every entity candidate discovered for a stream during Local EMD. It is incrementally updated with the local embeddings of a candidate's mentions. Next, a pooling operation conducted over all the local contextual embeddings of an entity candidate gives the `\textit{\textbf{global candidate embedding}}'. 
% \begin{equation} \scriptsize
%     global\_emb = \frac{1}{\left|\eta \right|}\sum_{j=1}^{\left|\eta \right|}a_{j}local\_emb_{\eta_{j}}
% \end{equation}
% \begin{equation} \scriptsize
%     a_{j}=W_{a}^{T}local\_emb_{j}+b_{a}
% \end{equation}
% \normalsize
% where the weights $W_{a}\epsilon \mathbb{R}^{d}$ and bias term $b_{a}$ are learnable parameters, $\eta$ denotes the set of mention-level local embeddings for a candidate, and $a_{j}$ is the weight of the local embedding ($local\_emb_{\eta_{j}} \epsilon \mathbb{R}^{d}$) of the j-th mention in $\eta$. 
% The embedding $global\_emb$ 
It is global in the sense that it aggregates all contextual possibilities in which a candidate appears in the stream, to generate a consensus representation.

The global candidate embeddings are then fed to a multi-layer network of feed-forward layers with ReLU activation function followed by a sigmoid output layer. We call this module the Entity Classifier. It is trained to determine the likelihood of a candidate being an entity. 
% For our implementation, we construct a fairly simple multi-layer network consisting of feed-forward layers with ReLU activation function followed by a sigmoid output layer. 
The sigmoid output gives the probability of a candidate being a true entity and is divided into three ranges, that we empirically determined from variation in the Classifier's entity detection performance over different values. Each range corresponds to a class label:\\
(1) $\alpha$: $\geq$0.55, candidate is confidently labelled as an \textbf{entity}.\\
(2) $\beta$: $\leq$0.4, candidate is confidently labelled as a \textbf{non-entity}.\\
(3) $\gamma$: $\epsilon$(0.4, 0.55), deemed \textbf{ambiguous}; requires more evidence downstream for confident labeling.

Note that a candidate's global embedding over mention variations is more reliable when its frequency of occurrence is high and is computed over more than just a few mentions (avoiding randomness). 
%This implies that the underlying syntactic trend, as inferred from the frequency features, is more reliable than what would appear due to sheer randomness. 
Consequently, the classifier performs better in distinguishing entities among more frequent candidates.

The classifier is supervised with the training performed using labelled global embedding records of entity candidates extracted from $D_5$ (see Table 1). Further details on the classifier training will be provided in Section \ref{sec:classifier-training}.
%We use a 80-20 training to validation split and train over 1000 epochs. We use Adam optimizer \cite{adam} with a fixed learning rate of 0.00001 and batch size of 32. %We also enable early stopping in our training. 
% We compute the task performance after each training epoch on the validation set, and select the best model checkpoint to compute the performance score on the test set. Here, we also apply early stopping when no improvement is observed on the validation loss after a patience of 25 continuous epochs.
% In future extensions we plan to convert this to a semi-supervised process that periodically expands the training set with test cases producing high-confidence outputs, where class labels remain unchanged over five consecutive processing cycles.

\textbf{Follow-up on case study on dataset $D_2$} : In Figure \ref{fig:coronavirus-EMD-followup}, it can be seen that all the entity mentions that were missed by Aguilar et al. \cite{aguilar-etal-WNUT17} and BERTweet \cite{bertweet} are discovered by the end of the EMD Globalizer pipeline.

\begin{table}[!ht]
\small
\centering
\vspace{20pt}
 \caption{Twitter Datasets}
 \label{twitter-dataset}
%  \vspace{-10pt}
 \begin{tabular}{ccccl}
   \toprule
 Dataset&Size&\#Topics&\#Hashtags&\#Entities\\
   \midrule
      $D_{1}$&1K&1&1&283\\
      $D_{2}$&2K&1&1&461\\
      $D_{3}$&3K&3&6&906\\
      $D_{4}$&6K&5&5&674\\
      $D_{5}$&38K&1&1&$\approx$7000\\
      WNUT17&1287&-&-&-\\
      BTC&9553&-&-&~\\
\bottomrule
\end{tabular}
 \vspace{-10pt}
\end{table}
\section{Experiments}
\label{sec:experiments}

We conducted extensive experiments to test the effectiveness of EMD Globalizer for entity mention detection in tweets. We used four existing EMD systems, including two deep EMD systems, for Local EMD. In each case, we evaluate the 
%enhancement in 
performance gain, when plugged into the framework. We implemented the framework in Python 3.8 and executed it on a NVIDIA Tesla T4 GPU on Google Colaboratory. %system with 3.60GHz Intel Core(TM) i7-4790 CPU and 8GB memory. 
Our %datasets and experimental codes are available on Github. 
% The Deep learning model was executed on a NVIDIA Tesla T4 GPU on Google Colaboratory. 
datasets and code used for experiments are available at Github.

\smallskip
\noindent
{\bf Datasets:}
\label{sec:dataset}
%topic clustering 
We use a combination of third-party datasets, along with the streaming datasets used in \cite{twics} that include crawled message streams from Twitter, for the purpose of evaluating the effectiveness of EMD Globalizer. %We follow the principle of focused crawling in data collection so that they reflect topical coherence (recall characteristic E-H1). 
The datasets are listed in Table \ref{twitter-dataset}. 
$D_1$-$D_4$ are \textbf{streaming datasets} that contain subsets of Twitter streams. The topics covered here are  
Politics, Sports, Entertainment, Science and Health, with ($D_2$) curated from a Covid-19 tweet stream. Having datasets directly covering a variety of tweet streams helps preserve their naturally occurring topic-specificity that often repeats a finite set of entities. This is exploited in Global EMD for collectively processing candidates, without making the analysis biased towards a particular topic. In real-world deployment, a topic classifier \cite{lee-twitter-topic-classifier} could precede an EMD tool launched for streams. %We seed the search with a set of top trending hashtags and iteratively enlarge the dataset by submitting queries with frequent n-grams and new hashtags. 

Other than the four streaming datasets $D_1$-$D_4$, two datasets popular for EMD benchmarking, WNUT17 \cite{wnut2017results} and BTC \cite{btc-dataset}, are also included in our evaluation. These are \textbf{non-streaming datasets} curated to accommodate a random sampling of tweets. Although they do not characterize the application setting for EMD Globalizer, we use them to gauge the framework's effectiveness against pre-established benchmarks. %$D_6$ and $D_7$ are not annotated. $D_7$ is also a third party dataset -- the 2011 Tweets Collection\footnote{\scriptsize \url{https://trec.nist.gov/data/tweets}} provided by NIST that was part of the TREC 2011 Microblog track \cite{TREC-overview}.

Also like \cite{twics}, we use dataset $D_5$, a collection of 38K tweets from a single tweet stream to generate entity candidates. The candidates are labelled as `entity'/`non-entity' and used to train the Entity classifier to learn optimal global embeddings and generate correct candidate labels for Global EMD.

\smallskip
\noindent
{\bf Performance Metrics:}
\label{sec:metric}
We use Precision ($P$), Recall ($R$) and F1-score to evaluate EMD effectiveness. EMD requires detection of all occurrences of entities in their various string forms within a dataset. It is captured in WNUT17 shared task \cite{wnut2017results} as \textbf{F1 (Surface)}. Here we simply call it F1. %Also note that we measure performance using 5-fold cross-validation with 4 folds to tune the Z-score threshold before candidate classification and one fold as the test data. We report the average performance numbers.
Our framework does not involve entity typing. So the evaluation here only includes EMD and not their type classification.
We also record execution times in seconds to check the run-time overhead for executing Local EMD systems within the framework. %Execution times are recorded in seconds.

\smallskip
\noindent
{\bf Local EMD Instantiations:~}
\label{sec:baselines}
We run our EMD framework with four different Local EMD instantiations (see Section \ref{local-EMD-instantiations}): %\textbf{CS} -- a syntactic heuristic based entity candidate tagger; 
1) \textbf{NP Chunker} -- a Chunking based Tagger that uses noun phrase chunking on Twitter dependency parser \cite{tweebo} to project entity candidates; 2) \textbf{Twitter NLP} \cite{Ritter} -- a CRF based Machine Learning model; 3) \textbf{Aguilar et al.} \cite{aguilar-etal-WNUT17} -- a Deep Learning architecture that won the WNUT 2017 \cite{wnut2017results} NER challenge; and finally, 4) \textbf{BERTweet} \cite{bertweet} -- a BERT language model trained on a large twitter corpus that we fine-tune using the WNUT2017 training data for the downstream EMD task.

\smallskip
\noindent
% \textcolor{blue}{
{\bf Baseline for testing Global EMD:~} We use the production version of the Document EMD system HIRE-NER \cite{hire-ner} as a baseline for testing Global EMD. We compare the performance of this system with EMD Globalizer on our Twitter datasets. HIRE-NER treats messages in a stream as composite content, much like a document.
% }

\smallskip
\noindent
{\bf Training Entity Phrase Embedder:~}
\label{sec:training-phrase-embedder}
When using Deep EMD systems for Local EMD, we employ an Entity Phrase Embedder in the Global EMD step to combine the contextual embeddings of a candidate's tokens provided by the deep EMD system into a unified local embedding for the entire candidate phrase. As mentioned earlier, the Entity Phrase Embedder is trained using the STS Benchmark (STS-b) data \cite{sts-benchmark}. This dataset contains 5749 sentence pairs in the training set and 1500 sentence pairs in the validation set. Every sentence pair is given a score between 0-5, indicating the semantic similarity between the sentences in the pair. To evaluate the Entity Phrase Embedder, we divide the integer scores by 5 to normalize them into a range of [0, 1], and then compare this with the Cosine similarities between the embeddings of the sentence pairs generated by the Entity Phrase Embedder. Here we use mean squared loss as the regression objective function to optimize training and estimate performance on the validation set.

We use Adam optimizer \cite{adam} with a fixed learning rate of 0.001 and batch size of 32. %We also enable early stopping in our training. 
We compute performance on the validation set after each training epoch, and save the best model checkpoint to execute test sets. Here, we also enforce early stopping when validation performance does not improve for 25 continuous epochs. Note that S-BERT \cite{sentence-BERT} tests the quality of sentence embeddings by employing them in downstream tasks like Paraphrase Mining and Natural Language Inference. Since we simply use the Entity Phrase Embedder to generate embeddings for candidate phrases, such detailed evaluation is not carried out here. 

We separately train the Entity Phrase Embedder for the two different variants of our framework with Aguilar et al. \cite{aguilar-etal-WNUT17} and BERTweet \cite{bertweet} as Local EMD systems. 
For Aguilar et al. the size of the candidate embeddings generated by the Entity Phrase Embedder is of 100 dimensions, the same as the system's output vectors. 
When trained with token embeddings from Aguilar et al., the best validation loss obtained is 0.185. 
For BERTweet, we tested EMD Globalizer with candidate embeddings of size 768 -- the size of the BERT encoder's output layer -- and 300. In our experiments, we obtained slightly better effectiveness with candidate embeddings of size 300 and hence we present those results in our evaluation in Table \ref{table:giant-table}. Nonetheless, these hyperparameters are easily customizable.
When trained with token embeddings from BERTweet, the best validation loss is 0.167.\\

\begin{table}[!hb]
\centering
\vspace{15pt}
\caption{Validation Performance of Entity Classifier}
\label{table:result-classifiers}
\vspace{1pt}
\begin{tabular}{|c|c|c|c|}
\hline
Local EMD      & \begin{tabular}[c]{@{}c@{}}Local EMD\\ System Type\end{tabular} & \begin{tabular}[c]{@{}c@{}}Entity\\ Embedding\\ Size\end{tabular} & \begin{tabular}[c]{@{}c@{}}Validation\\ F1\end{tabular} \\ \hline
NP Chunker     & CRF Chunker                                                     & 6+1                                                                 & 0.936                                                    \\ \hline
TwitterNLP     & CRF EMD Tagger                                                  & 6+1                                                                 & 0.936                                                    \\ \hline
Aguilar et al. & BiLSTM-CNN-CRF                                                  & 100+1                                                               & 0.908                                                    \\ \hline
BERTweet       & BERT-FFNN                                                        & 300+1                                                               & 0.941                                                    \\ \hline
\end{tabular}
\end{table}

\begin{table*}[!ht]
\centering
% \footnotesize
\vspace{10pt}
\caption{Effectiveness and Execution Time (in seconds) with EMD Globalizer}
\label{table:giant-table}
\vspace{1pt}
\begin{tabular}{|c|c|c|c|c|c|cccc|c|c|}
\hline
\multirow{2}{*}{Dataset}         & \multicolumn{5}{c|}{Local EMD}                                                                                                        & \multicolumn{4}{c|}{Global EMD}                                                                                                                & \multirow{2}{*}{F1 Gain} & \multirow{2}{*}{\begin{tabular}[c]{@{}c@{}}Time\\ Overhead\end{tabular}} \\ \cline{2-10}
                                 & \begin{tabular}[c]{@{}c@{}}System\\ Name\end{tabular} & P    & R    & F1   & \begin{tabular}[c]{@{}c@{}}Execution\\ Time\end{tabular} & \multicolumn{1}{c|}{P}    & \multicolumn{1}{c|}{R}    & \multicolumn{1}{c|}{F1}    & \begin{tabular}[c]{@{}c@{}}Execution\\ Time\end{tabular} &                          &                                                                          \\ \hline
\multirow{4}{*}{\textbf{D1}}     & NP Chunker                                            & 0.3  & 0.58 & 0.4  & 100.4                                                    & \multicolumn{1}{c|}{0.81} & \multicolumn{1}{c|}{0.63} & \multicolumn{1}{c|}{0.71}  & 101.6                                                    & 77.5\%                   & 1.2                                                                      \\ \cline{2-12} 
                                 & TwitterNLP                                            & 0.65 & 0.47 & 0.55 & 7.07                                                     & \multicolumn{1}{c|}{0.8}  & \multicolumn{1}{c|}{0.66} & \multicolumn{1}{c|}{0.72}  & 8.03                                                     & 36.4\%                   & 0.96                                                                     \\ \cline{2-12} 
                                 & Aguilar et al.                                        & 0.76 & 0.55 & 0.64 & 124.8                                                    & \multicolumn{1}{c|}{0.87} & \multicolumn{1}{c|}{0.66} & \multicolumn{1}{c|}{0.75}  & 126.07                                                   & 17.3\%                   & 1.27                                                                     \\ \cline{2-12} 
                                 & BERTweet                                              & 0.66 & 0.49 & 0.56 & 33.16                                                    & \multicolumn{1}{c|}{0.84} & \multicolumn{1}{c|}{0.66} & \multicolumn{1}{c|}{0.74}  & 34.32                                                    & 32.1\%                   & 1.16                                                                     \\ \hline
\multirow{4}{*}{\textbf{D2}}     & NP Chunker                                            & 0.40 & 0.47 & 0.43 & 123.62                                                   & \multicolumn{1}{c|}{0.59} & \multicolumn{1}{c|}{0.62} & \multicolumn{1}{c|}{0.60}  & 125.71                                                   & 39.5\%                   & 2.09                                                                     \\ \cline{2-12} 
                                 & TwitterNLP                                            & 0.33 & 0.52 & 0.41 & 18.91                                                    & \multicolumn{1}{c|}{0.71} & \multicolumn{1}{c|}{0.55} & \multicolumn{1}{c|}{0.62}  & 20.57                                                    & 51.2\%                   & 1.66                                                                     \\ \cline{2-12} 
                                 & Aguilar et al.                                        & 0.63 & 0.57 & 0.60 & 296                                                      & \multicolumn{1}{c|}{0.69} & \multicolumn{1}{c|}{0.67} & \multicolumn{1}{c|}{0.68}  & 297.7                                                    & 13.3\%                   & 1.7                                                                      \\ \cline{2-12} 
                                 & BERTweet                                              & 0.56 & 0.51 & 0.53 & 40.23                                                    & \multicolumn{1}{c|}{0.65} & \multicolumn{1}{c|}{0.64} & \multicolumn{1}{c|}{0.64}  & 42.58                                                    & 20.8\%                   & 2.35                                                                     \\ \hline
\multirow{4}{*}{\textbf{D3}}     & NP Chunker                                            & 0.59 & 0.54 & 0.56 & 175.3                                                    & \multicolumn{1}{c|}{0.71} & \multicolumn{1}{c|}{0.66} & \multicolumn{1}{c|}{0.68}  & 177.9                                                    & 21.4\%                   & 2.6                                                                      \\ \cline{2-12} 
                                 & TwitterNLP                                            & 0.75 & 0.64 & 0.69 & 15.1                                                     & \multicolumn{1}{c|}{0.88} & \multicolumn{1}{c|}{0.71} & \multicolumn{1}{c|}{0.78}  & 18                                                       & 13.04\%                  & 2.9                                                                      \\ \cline{2-12} 
                                 & Aguilar et al.                                        & 0.77 & 0.64 & 0.70 & 298.2                                                    & \multicolumn{1}{c|}{0.82} & \multicolumn{1}{c|}{0.77} & \multicolumn{1}{c|}{0.794} & 301.34                                                   & 13.6\%                   & 3.14                                                                     \\ \cline{2-12} 
                                 & BERTweet                                              & 0.77 & 0.63 & 0.69 & 58.6                                                     & \multicolumn{1}{c|}{0.83} & \multicolumn{1}{c|}{0.82} & \multicolumn{1}{c|}{0.83}  & 62.18                                                    & 20.3\%                   & 3.58                                                                     \\ \hline
\multirow{4}{*}{\textbf{D4}}     & NP Chunker                                            & 0.47 & 0.59 & 0.52 & 551.3                                                    & \multicolumn{1}{c|}{0.83} & \multicolumn{1}{c|}{0.73} & \multicolumn{1}{c|}{0.77}  & 556.7                                                    & 48.1\%                   & 5.4                                                                      \\ \cline{2-12} 
                                 & TwitterNLP                                            & 0.67 & 0.41 & 0.52 & 35.24                                                    & \multicolumn{1}{c|}{0.89} & \multicolumn{1}{c|}{0.64} & \multicolumn{1}{c|}{0.74}  & 41.06                                                    & 42.3\%                   & 5.82                                                                     \\ \cline{2-12} 
                                 & Aguilar et al.                                        & 0.82 & 0.61 & 0.69 & 588.24                                                   & \multicolumn{1}{c|}{0.88} & \multicolumn{1}{c|}{0.75} & \multicolumn{1}{c|}{0.81}  & 594.22                                                   & 17.4\%                   & 5.98                                                                     \\ \cline{2-12} 
                                 & BERTweet                                              & 0.69 & 0.58 & 0.62 & 230.75                                                   & \multicolumn{1}{c|}{0.81} & \multicolumn{1}{c|}{0.76} & \multicolumn{1}{c|}{0.78}  & 237.53                                                   & 26.1\%                   & 6.78                                                                     \\ \hline
\multirow{4}{*}{\textbf{WNUT17}} & NP Chunker                                            & 0.42 & 0.35 & 0.39 & 121.22                                                   & \multicolumn{1}{c|}{0.63} & \multicolumn{1}{c|}{0.35} & \multicolumn{1}{c|}{0.44}  & 123.56                                                   & 12.8\%                   & 2.34                                                                     \\ \cline{2-12} 
                                 & TwitterNLP                                            & 0.35 & 0.42 & 0.39 & 14.25                                                    & \multicolumn{1}{c|}{0.65} & \multicolumn{1}{c|}{0.52} & \multicolumn{1}{c|}{0.58}  & 16.72                                                    & 48.7\%                   & 2.47                                                                     \\ \cline{2-12} 
                                 & Aguilar et al.                                        & 0.68 & 0.47 & 0.56 & 229.32                                                   & \multicolumn{1}{c|}{0.72} & \multicolumn{1}{c|}{0.5}  & \multicolumn{1}{c|}{0.59}  & 231.04                                                   & 5.4\%                    & 1.72                                                                     \\ \cline{2-12} 
                                 & BERTweet                                              & 0.61 & 0.43 & 0.51 & 24.40                                                    & \multicolumn{1}{c|}{0.73} & \multicolumn{1}{c|}{0.48} & \multicolumn{1}{c|}{0.58}  & 26.15                                                    & 13.7\%                   & 1.75                                                                     \\ \hline
\multirow{4}{*}{\textbf{BTC}}    & NP Chunker                                            & 0.46 & 0.51 & 0.48 & 627.98                                                   & \multicolumn{1}{c|}{0.66} & \multicolumn{1}{c|}{0.52} & \multicolumn{1}{c|}{0.58}  & 642.02                                                   & 20.8\%                   & 14.04                                                                    \\ \cline{2-12} 
                                 & TwitterNLP                                            & 0.69 & 0.43 & 0.53 & 77.15                                                    & \multicolumn{1}{c|}{0.74} & \multicolumn{1}{c|}{0.45} & \multicolumn{1}{c|}{0.56}  & 87.8                                                     & 5.7\%                    & 10.65                                                                    \\ \cline{2-12} 
                                 & Aguilar et al.                                        & 0.75 & 0.56 & 0.64 & 685.36                                                   & \multicolumn{1}{c|}{0.77} & \multicolumn{1}{c|}{0.59} & \multicolumn{1}{c|}{0.67}  & 695.56                                                   & 4.7\%                    & 10.2                                                                     \\ \cline{2-12} 
                                 & BERTweet                                              & 0.63 & 0.50 & 0.56 & 193.8                                                    & \multicolumn{1}{c|}{0.69} & \multicolumn{1}{c|}{0.58} & \multicolumn{1}{c|}{0.63}  & 204.49                                                   & 12.5\%                   & 10.69                                                                    \\ \hline
\end{tabular}
\end{table*}

\smallskip
\noindent
{\bf Training Entity Classifier:~}
\label{sec:classifier-training}
We train the Entity Classifier everytime we initialize a variant of the framework with a different Local EMD instantiation. The `+1' in the column for Entity Embedding Size (see Table \ref{table:result-classifiers}) indicates that we also add length of the candidate string as an additional feature, along with the global entity embedding. % the choice of classifier algorithm does not create much significant impact on performance for different Local EMDs. But overall, the SVM classifier performs best among the three variants across our datasets. Therefore, we use SVM as the default classifier for other experiments.
We use a 80-20 training to validation split and train over 1000 epochs. We use Adam optimizer \cite{adam} with a fixed learning rate of 0.0015 and batch size of 128. %We also enable early stopping in our training. 
We compute the task performance after each training epoch on the validation set, and select the best model checkpoint to compute the performance score on the test set. Here, we also apply early stopping when no improvement is observed on the validation loss after 20 continuous epochs. These validation performance obtained when training the Entity Classifier for different variants of our framework are compiled in Table \ref{table:result-classifiers}.
%In future work we plan to convert this to a semi-supervised process that periodically expands the training set with test cases producing high-confidence outputs.\\ % where class labels remain unchanged over several processing cycles.
\begin{table}[!htb]
\centering
\vspace{10pt}
\caption{Effectiveness of Global EMD systems}
\label{table:globalizer-vs-docEMD}
\vspace{1pt}
%\textcolor{blue}{
\begin{tabular}{|c|c|l|l|l|}
\hline
\multicolumn{1}{|l|}{Dataset} & \multicolumn{1}{l|}{Global EMD System} & P    & R    & F1   \\ \hline
\multirow{2}{*}{D1}           & EMD Globalizer                         & 0.87  & 0.66 & 0.75 \\ \cline{2-5} 
                              & HIRE-NER                               & 0.65 & 0.62 & 0.63 \\ \hline 
                            %   & DocL-NER                               & 0.61 & 0.64 & 0.62 \\ \hline
\multirow{2}{*}{D2}           & EMD Globalizer                         & 0.69 & 0.67 & 0.68 \\ \cline{2-5} 
                              & HIRE-NER                               & 0.46 & 0.56 & 0.51 \\ \hline
                            %   & DocL-NER                               & 0.52 & 0.62 & 0.56 \\ \hline
\multirow{2}{*}{D3}           & EMD Globalizer                         & 0.82 & 0.77 & 0.79 \\ \cline{2-5} 
                              & HIRE-NER                               & 0.75 & 0.73 & 0.74 \\ \hline 
                            %   & DocL-NER                               & 0.74 & 0.72 & 0.73 \\ \hline
\multirow{2}{*}{D4}           & EMD Globalizer                         & 0.88 & 0.75 & 0.81 \\ \cline{2-5} 
                              & HIRE-NER                               & 0.58 & 0.68 & 0.61 \\ \hline 
                            %   & DocL-NER                               & 0.53 & 0.68 & 0.59 \\ \hline
\multirow{2}{*}{WNUT}         & EMD Globalizer                         & 0.72 & 0.5  & 0.59 \\ \cline{2-5} 
                              & HIRE-NER                               & 0.5  & 0.49  & 0.5  \\ \hline 
                            %   & DocL-NER                               & 0.49 & 0.47 & 0.48 \\ \hline
\multirow{2}{*}{BTC}          & EMD Globalizer                         & 0.77 & 0.59 & 0.67 \\ \cline{2-5} 
                              & HIRE-NER                               & 0.6  & 0.49 & 0.54 \\ \hline 
                            %   & DocL-NER                               & 0.56 & 0.48 & 0.52 \\ \hline
\end{tabular}
% }
%\vspace{-4pt}
\end{table}

\subsection{Evaluating EMD Globalizer}
We first evaluate EMD Globalizer on its primary objective of boosting EMD for the Local EMD systems. 
To this end, we test EMD Globalizer with four different Local EMD instantiations. In each case we check the improvement that is achieved from the Local EMD's initial F1 score by executing it with the rest of the framework. Table \ref{table:giant-table} summarizes these results along with the execution time overhead brought in for each Local EMD system post plugin. We also compare the best performing EMD Globalizer variant to a state-of-the-art Document EMD system to understand how effectively global information is mined in each case and utilized for EMD.
%Table \ref{table:giant-table} shows the improvements our framework produced in EMD performance of different Local EMD instantiations compared to their standalone versions. We analyze both the improvement in terms of effectiveness and the execution time overhead brought in by the rest of the framework, post plugin. The performance metrics are computed. 

\smallskip
\noindent
\textbf{Local EMD Performance:} The columns under `Local EMD' in Table \ref{table:giant-table} show the EMD performances of each of the four Local EMD systems, along with their computation times. %Considering that a system being used for Local EMD happens to be the only component in the framework at this point, it also indicates the system's standalone performance. 

\smallskip
\noindent
\textbf{Performance improvement with EMD Globalizer: } The columns in Table \ref{table:giant-table} under `Global EMD' show the EMD performance once the Global EMD components have been run on top of a Local EMD system and the total run-time at the end of its execution. %i.e. after both Local EMD and Global EMD components have completed processing tweets. 
Comparing the F1 of a Local EMD system with that of its Global EMD counterpart 
%For a particular Local EMD system, the corresponding EMD performance at the end of Global EMD's execution 
gives the improvement achieved by the framework. The difference in execution times at the end of the two stages gives the additional time required by Global EMD.
%for this.

Global EMD is able to make considerable improvement in performance with only minor execution time overhead across all datasets.
%that we have used for testing. 
For each Local EMD system, we compute the percentage gain in F1 score across all datasets. For time overhead, we calculate the absolute increment in the execution time of a system (in seconds) when passed through EMD Globalizer. %, as well as the percentage gain in execution time (numbers within braces)
This provides better context. %For example, for CS which happens to be the least computationally expensive tagger, the time overheads are relatively higher than other systems, across the different datasets. 
As is the case of every Local EMD system, the absolute overhead incurred by injecting it into the framework is only a few additional seconds. For computationally expensive EMD systems that already have higher execution times, the time overhead is nominal, thereby making the performance gain obtained all that more significant. In summary, the average performance gain across all datasets and all local EMD systems is 25.61\%. The average individual performance gains for the four Local EMD systems are: a) 36.69\% for NP Chunker, b) 31.06\% for TwitterNLP, c) 11.91\% for Aguilar et al., and, d) 20.66\% for BERTweet. 
% That gives an average overall improvement of 25.54\% across all datasets and Local EMD instantiations.
Note that two types of datasets are used in our evaluation, one is the streaming datasets -- the application setting for which EMD Globalizer was originally designed, and the other is non-streaming datasets -- popularly used for EMD benchmarking. EMD Globalizer yields different improvements over these dataset types as discussed below. 

\smallskip
\noindent
\textbf{Improvement on Streaming Datasets:} For datasets $D_1$-$D_4$ that retain the inherent properties of Twitter streams, EMD Globalizer yields an average F1 gain of 30.29\% across all Local EMD systems. For individual Local EMD instantiations, the average F1 gains are: a) 46.63\% for NP Chunker, b) 34.36\% for TwitterNLP, c) 15.36\% for Aguilar et al., and, d) 24.82\% for BERTweet.

\smallskip
\noindent
\textbf{Improvement on Non-Streaming Datasets:} For datasets WNUT17 and BTC, there is no adherence to specific Twitter streams %and it's message generation is maintained. Rather, these datasets deliberately curate a 
but rather a random sampling off the Twittersphere, avoiding the latter's tendency to repeat entity mentions within streams. However, EMD Globalizer is still able to improve effectiveness for its Local EMD instantiations, albeit to a less significant degree than streaming datasets. In this case, the average F1 gain across all Local EMD systems is 15.53\%. For individual Local EMD instantiations, the average F1 gains are: a) 16.82\% for NP Chunker, b) 24.47\% for TwitterNLP, c) 5.04\% for Aguilar et al., and, d) 12.22\% for BERTweet.
% \begin{table*}[!htb]
% \centering
% \vspace{15pt}
% \caption{Performance Gain vs Time Overhead in seconds (T.O.)}
% \label{table:result-summary}
% \vspace{1pt}
% \begin{tabular}{|c|c|c|c|c|c|c|c|c|c|c|c|c|}
% \hline
% Dataset        & \multicolumn{2}{c|}{D\textsubscript{1}} & \multicolumn{2}{c|}{D\textsubscript{2}} & \multicolumn{2}{c|}{D\textsubscript{3}} & \multicolumn{2}{c|}{D\textsubscript{4}} & \multicolumn{2}{c|}{WNUT17} & \multicolumn{2}{c|}{BTC} \\ \hline
% Local EMD & F1 Gain & T.O. & F1 Gain & T.O. & F1 Gain & T.O. & F1 Gain & T.O. & F1 Gain  & T.O.  & F1 Gain  & T.O.  \\ \hline
% % CS          & 19.1\%     & 1     & 46.5\%     & 1.53     & 29.5\%   & 2.6      & 30.2\%   & 2.6   & 16.3\%    & 1.42        \\ \hline
% NP Chunker     & 77.5\%   & 1.2        & 39.5\%     & 2.09     & 21.4\%   & 2.6     & 48.1\%   & 5.4      & 12.8\%    & 2.34       & 20.8\%    & 14.04  \\ \hline
% TwitterNLP     & 36.4\%     & 0.96     & 51.2\%     & 1.66     & 13.04\%   & 2.9       & 42.3\%   &5.82       & 48.7\%    & 2.47     & 5.7\%    & 10.65  \\ \hline
% Aguilar et al. & 17.3\%   & 1.27      & 13.3\%     & 1.7     & 13.6\%   & 3.14     & 17.4\%   & 5.98      & 5.4\%    & 1.72       & 4.7\%    & 10.2  \\ \hline
% BERTweet & 32.1\%   & 1.16      & 20.8\%     & 2.35     & 20.3\%   & 3.58     & 26.1\%   & 6.78      & 13.7\%    & 1.75        & 10.8\%    & 10.69  \\ \hline
% \end{tabular}
% \end{table*}

\smallskip
\noindent
% \textcolor{blue}{
\textbf{Comparison with Document EMD method on Global EMD:} We compare the performance of %two document-level EMD systems, HIRE-NER \cite{hire-ner} and DocL-NER \cite{docL-ner} with 
Aguilar et al. instantiated EMD Globalizer and HIRE-NER \cite{hire-ner} (both BiLSTM architectures) on all the annotated datasets in Table \ref{table:globalizer-vs-docEMD}. Here we test how effectively global information is captured in each system when performing EMD.
As evident from Table \ref{table:globalizer-vs-docEMD}, EMD Globalizer consistently outperforms HIRE-NER across all datasets, especially attaining higher precision. HIRE-NER %and DocL-NER 
simultaneously updates global features in the memory structure and appends them with local embeddings to infer final output labels of tokens in a sentence. Adding non-local contextual information inevitably introduces noise which can interfere with the decoder's inference of output labels. Distinct from this, we limit the curation of global contextual representations only for entity candidates. First the local context suggests the entity candidates situated within a sentence. The local contextual embeddings of the various candidate mentions are then aggregated in the memory structure to generate global candidate embeddings. Using this the entity classifier is able to better separate 
%legitimate 
true entities from noisy candidates.% under the assumption that the more we encounter a candidate, the better we learn to represent it.
%All systems show reduced effectiveness for the non-streaming datasets where the token recurrence of streams or documents is rare, making it difficult to be utilized to aggregate global features.
% }

\subsection{Ablation Study on Framework Components}
\begin{figure}[!ht]
% \vspace{-10pt}
\centering
\includegraphics[scale=0.37]{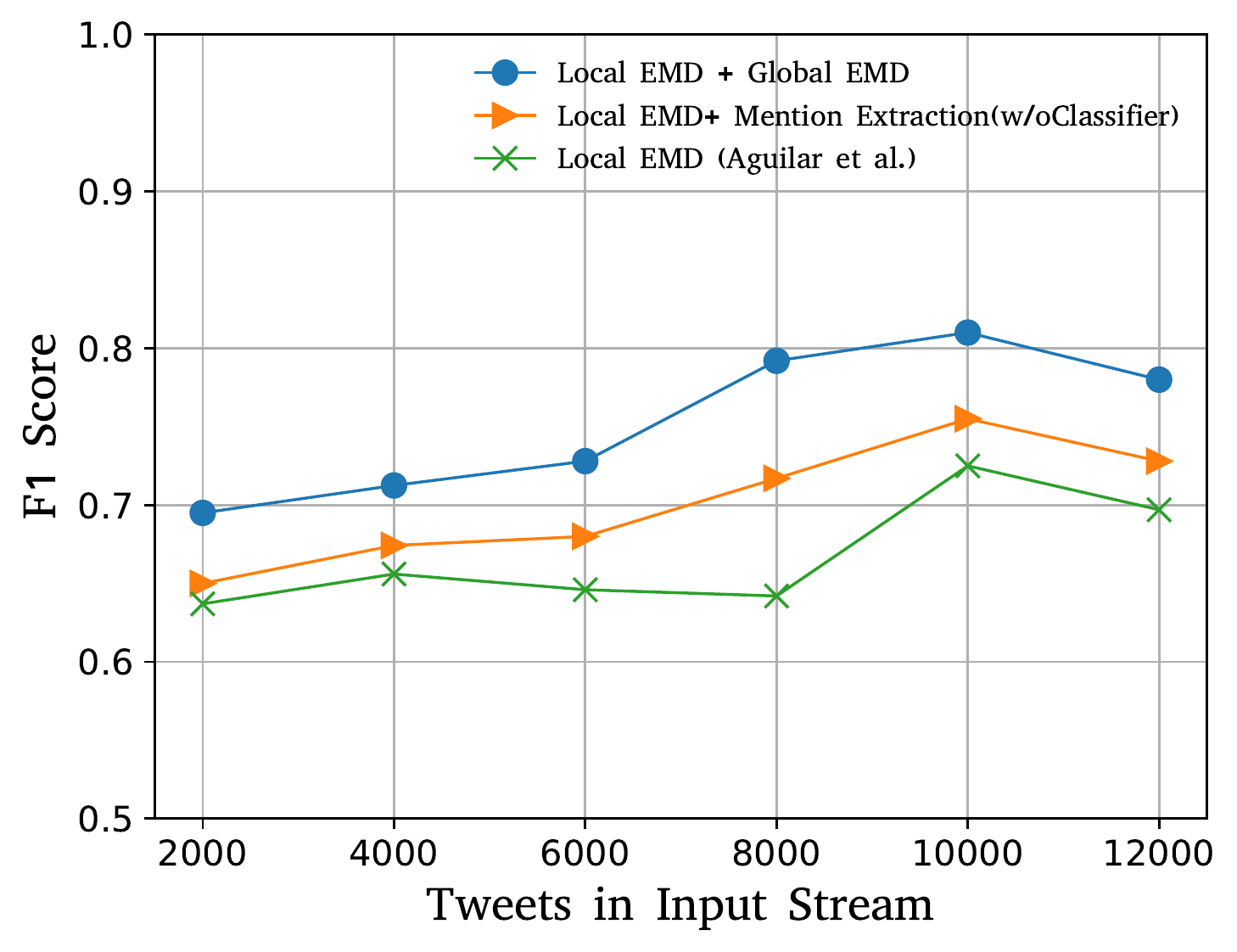}\par
%\vspace{-10pt}
\caption{Impact of Components on Performance}\label{fig:components}
\vspace{0.2cm}
\end{figure}

While it is evident that the proposed EMD framework is capable of enhancing performance for its various Local EMD instantiations, we wanted to take a closer look at how the individual framework components contribute towards the EMD overall performance. To this end, we execute the framework with Aguilar et al. \cite{aguilar-etal-WNUT17} as the Local EMD system in it as this instantiation is the best performer among the local EMD systems used in this paper. Here, we use the entire collection of annotated streaming datasets ($D_1$-$D_4$) as the test set.
Figure \ref{fig:components} shows the improvement in performance as individual system components are added. From bottom to top, the first curve (with only Local EMD) reports the weakest performance proving the limitations of the standalone system in capturing all the entity mention variations within the stream. %The topmost curve from bottom is a once-and-done version of the framework while the topmost one is the full Rescan enabled framework. %The performance gain makes it clear why systems like a CS based tagger that prioritized computational efficiency over sophistication, surely benefits by having a selective Rescan strategy in place. We will also establish the fact that the memory overhead for Rescan is fairly minute, even for larger datasets i.e. busy streams where larger memory footprints are not desirable.
The middle curve is the EMD performance we get just by following up the entity (candidate) extraction by Local EMD with the mention extraction process that simply adds missed mentions of candidates detected as likely entities by the local EMD system.
The topmost curve is the performance yielded at the end of run of the entire framework with Aguilar et al. \cite{aguilar-etal-WNUT17}. Aguilar et al. \cite{aguilar-etal-WNUT17} is a very competitive EMD system -- the best among the four (local) systems we evaluated in this paper. Even for such a system, EMD Globalizer is still able to significantly improve on its EMD effectiveness over the streaming datasets. Following up its execution with just the candidate mention extraction process gives a modest improvement of 5.06\%. This simply recovers missed mentions of candidates identified as entities elsewhere in the stream.
The focus here is mainly on improving the recall by yielding more consistent mention detection across tweets. With EMD Globalizer however, the average overall improvement across all streaming datasets is 15.36\%. This is because with all components of Global EMD in place, candidates suggested by Aguilar et al. are further verified and false positives are removed. Table \ref{table:giant-table} shows that both precision and recall are improved with the induction of Global EMD.

\subsection{Error Analysis}
% \textcolor{blue}{
%Here we take a closer look into issues that impact the EMD Globalizer's ability to find entity mentions.}
%
Though EMD Globalizer improves over Local EMDs, it is not perfect. We give here an analysis of its errors.%}

% \textcolor{blue}{
1) %Local EMD fails to find at least one mentions of an entity
%Since Global EMD receives entity candidates from Local EMD, it is imperative that the latter finds at least one mention of a legitimate entity to register it as a candidate in the CTrie. Then the global step can discover other missed mentions. However 
If Local EMD misses every mention of an entity, the entity will not be added as a candidate to the CTrie and will also go undetected in the global phase. As a result, all mentions of the entity will not be detected by the proposed framework. Of the 11412 mentions in our streaming datasets from 2306 unique entities, the BERTweet instantiated EMD Globalizer failed to find 3008 (26.35\%) mentions of 1018 entities that are entirely missed by the BERTweet system.
% }

% \textcolor{blue}{
2) If Global EMD mislabels a candidate that happens to be an entity, then all of its mentions will be left out of the final output. This would include the mentions that the Local EMD did correctly find at first.  But more importantly, a false negative from Entity Classifier would hinder EMD Globalizer's objective of recovering mentions of the entity that the Local EMD missed. However, in our experience, it is rare that an entity found by Local EMD is missed at the global step. 
Of the 11412 entity mentions in our streaming datasets, BERTweet instantiated EMD Globalizer missed only 469 mentions (4.1\%) due to the mislabeling of 81 entities as false negatives by the Entity Classifier. 
%Of the 11412 entity mentions in our streaming datasets, there were only 129 (1.13\%) mentions that Local EMD found but Global EMD was unable to produce in its final output. More importantly, a false negative from the classifier would hinder the EMD Globalizer's objective of improving the Local EMD's findings by recovering mentions of the entity that the latter missed.
% }

\begin{figure}[!t]
\vspace{-2pt}
\centering
\includegraphics[height=1.55in,width=2.25in]{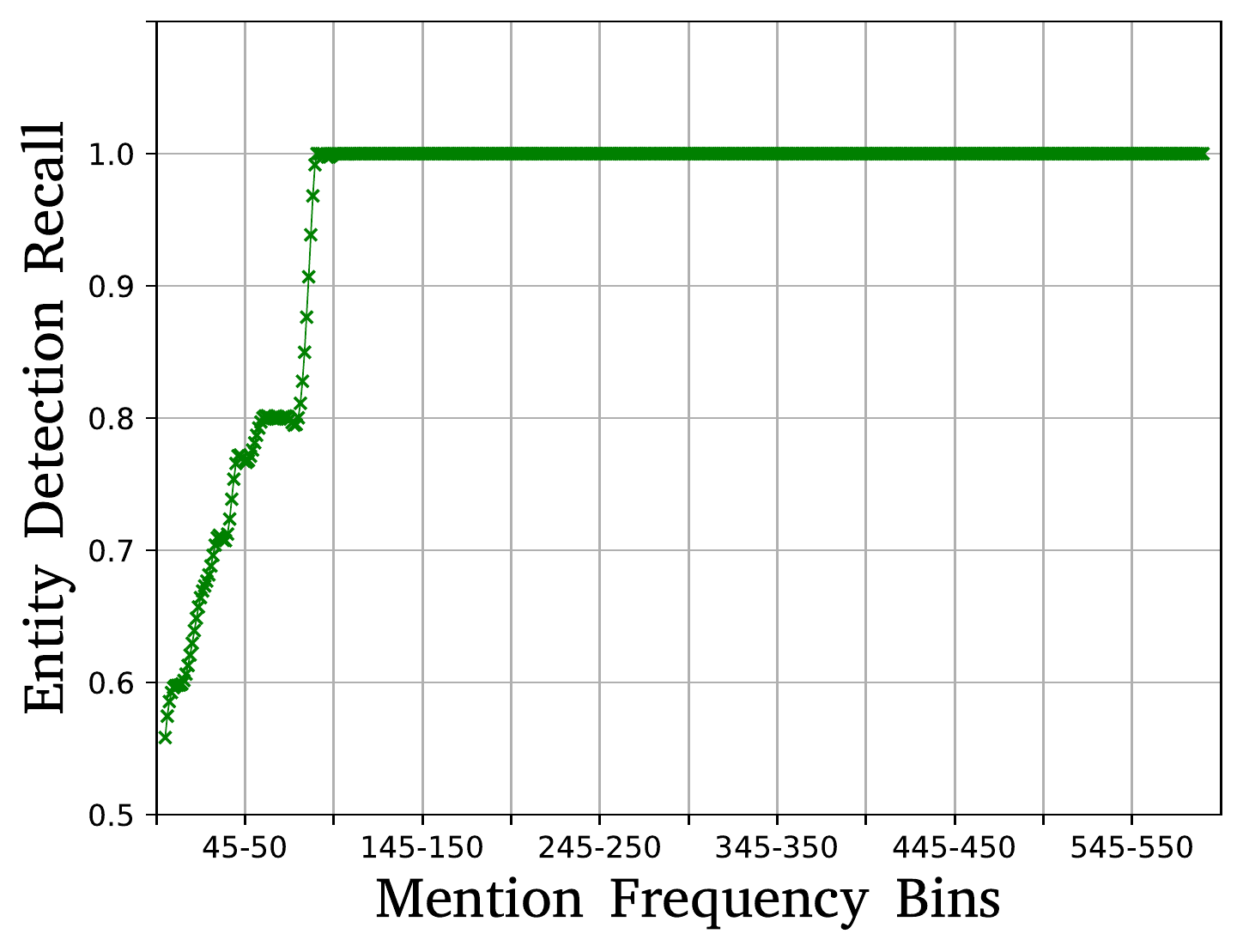}
%[height=2.5in, width=2.5in]
\caption{Impact of Frequency on Detecting Entities}\label{fig:frequency-wise-recall}
% \vspace{2pt}
\end{figure}

% \textcolor{blue}{
3) Handling of long-tailed entities: To better understand the false negatives yielded by Global EMD, we take a look at how the Entity Classifier's performance changes as more mentions of an entity are found in a stream. Figure \ref{fig:frequency-wise-recall} shows that it is consistently able to detect high-frequency entities from the streaming datasets. We group entities of different mention frequency in bins of width 5 and track the classifier's recall in detecting them.
For infrequent entities, the recall is modest-- around 56\% for entities with 5 or less mentions. 
But it increases quickly with mention frequency and most frequent entities are correctly labelled. This ensures that their mentions are also included in the final EMD Globalizer output. 
This confirms our initial intuition that as more mentions of an entity are found, better global contextual embeddings can be learned, leading to correct classification. Although long-tailed entities are a common issue in EMD, EMD Globalizer can still rectify the mislabeling of many such entities by collecting more instances further downstream.
% }

% \textcolor{blue}{
EMD Globalizer aims to correct the often irregular detection of mentions of the same entity and we expect an improvement in EMD Recall from the Local EMD step. As seen in Table \ref{table:giant-table}, the EMD Globalizer is able to improve Aguilar et al.'s recall on the streaming datasets by 20.2\%  and BERTweet's recall by 30.3\% on average. Moreover, EMD Globalizer also does a good job of filtering noise when aggregating non-local information. As observed in the Global EMD baseline, this issue can affect the EMD precision. EMD Globalizer not only improves the precision yielded by Local EMD, it also yields fewer false positives than HIRE-NER %and DocL-NER 
(in Table \ref{table:globalizer-vs-docEMD}) on all datasets. From Table \ref{table:giant-table}, EMD Globalizer is able to improve Aguilar et al.'s precision by 10.1\% on average and BERTweet's by 17.1\% for the streaming datasets.
% }

%\vspace{-5pt}
\subsection{Discussions}
We make some additional interesting observations from Table \ref{table:giant-table}. Here we summarize their implications for EMD:
\begin{itemize}[leftmargin=*]
    \item \textbf{Streaming vs Non-streaming Datasets}: Even though EMD Globalizer was conceptualized with the streaming setting in mind, all Local EMD instantiations, including Aguilar et al. -- the erstwhile topper of WNUT17 NER challenge \cite{wnut2017results} -- experience performance gain within the framework, on the two widely-used non-streaming datasets. This further validates the power of this framework in maximizing performance across different dataset types. Nonetheless, the framework is designed mainly to improve EMD performance on message streams. The idea of generating global contextual embeddings guided by a candidate mention extraction process specifically relies on the recurrence of entities across messages -- a phenomenon more typical of social media message streams. Hence for streaming datasets, higher EMD performance is achieved with EMD Globalizer.
    % \item The choice of classifier in Global EMD does not seem to make a significant difference in the Global EMD, with the performance being consistent across different classification algorithms. This certifies that the performance gains mostly come from Forward Scan and Rescan.
    \item \textbf{Design Flexibility}: We deliberately decouple the Local EMD step -- which can be any existing EMD tool -- from the rest of the EMD Globalizer framework. The advantage here is that the Local EMD tool can be inserted as is without any algorithmic modification. Also, depending on the type of Local EMD tool, the individual components of Global EMD are separately customizable. 
    \item \textbf{Improvement for low-performing EMD systems}: Local EMD variants like the NP Chunker that initially produced sporadic and relatively lower effectiveness, also yield competitive performance when aided by the framework.
    \item \textbf{New state-of-the-art record on existing benchmarks}: Aguilar et al. outperformed its original performance of the WNUT17 challenge when executed within the framework, and, along with BERTweet and TwitterNLP, all record much improved results on WNUT17. BERTweet also improves upon its performance on WNUT17 as reported in \cite{bertweet} when executed within the framework.
    \item \textbf{Time Overhead}: The time overhead brought by inserting a Local EMD system into EMD Globalizer is a fraction of its standalone execution time and depends on the input/stream size. %For EMD systems that are computationally expensive, the overhead is miniscule. For example, Aguilar et al. incurred the least overhead averaging at approximately 1\% of its original execution time, when incorporated into the EMD Globalizer framework. On the other hand TwitterNLP is more efficient in its standalone execution and the average time overhead for TwitterNLP with the framework is approximately 14.8\% of its execution time. 
    The absolute overhead is a few additional seconds.
\end{itemize}

\vspace{-8pt}
\section{Conclusions}
\label{sec:conclusion}
In this paper, we presented EMD Globalizer -- a novel two-phase EMD framework designed to address the limitations of existing EMD systems when executed on microblog streams and improve their effectiveness. %EMD Globalizer is powered by a novel two-phase framework. 
Although EMD Globalizer is itself not a standalone EMD system, it is capable of significantly improving the EMD effectiveness of existing EMD systems that perform EMD on microblogs individually. In this paper, we tested EMD Globalizer with four existing EMD systems of various types (two deep EMD systems and two non-deep EMD systems) on both streaming and non-streaming datasets. Remarkable improvement on effectiveness (based on F1-measure) is achieved for each of these systems, averaging over 25\% across all the four EMD systems. The improvement is even more remarkable for streaming datasets, averaging over 30\% across the four systems. EMD Globalizer is specifically designed for streaming datasets. These improvements were achieved with only a small execution time overhead. 

%It begins with an existing EMD algorithm that aims at better generalization of the EMD problem by analyzing the immediate context of a message. We call this the Local EMD step. This is followed by the Global EMD step that begins by taking the outputs of Local EMD as entity candidates. 

A big reason that EMD Globalizer can achieve these improvements is its ability to aggregate 
%The purpose of Global EMD is to extract every mention of an entity candidate and collectively analyze the various contexts in which they appear. Everytime a mention of an entity candidate is encountered, the Global EMD module generates a local candidate embedding from the immediate context of the mention. 
local candidate embeddings (contexts of local entity candidates) into a global candidate embedding. It is global in the sense that it considers all the contexts in which a candidate appears within a stream and derives a consensus representation. The global candidate embedding is then used to determine the likelihood of an candidate being an entity. Candidates labelled as entities find their mentions in the final output.
%as input and uses the input to determine true positives and true negatives of EMs and to find all entity mentions of determined true positives from both the past and future messages of a targeted microblog stream. 
%In our experiments EMD Globalizer was tested with four different Local EMD instantiations on a variety of Twitter datasets. 
%EMD Globalizer was able to improve the effectiveness (based on F1-measure) of these Local EMD instantiations remarkably, across both streaming and non-streaming datasets, averaging more than 25\%. 

An encouraging result from our experiments was that EMD Globalizer also achieved good improvement for standard non-streaming datasets, averaging over 15\%. Based on this, we believe that EMD Globalizer is a powerful tool that could be %utilized to improve existing EMD systems 
applied for different EMD application settings, not just microblog streams. In future work, we aim to expand the idea of collective processing for the entire NER pipeline.

\setlength{\parindent}{0pt}
\paragraph*{\textbf{Acknowledgements}} This work was supported in part by U.S. National Science Foundation BIGDATA 1546480 and 1546441 grants, and a gift from NVIDIA Corporation.

\bibliographystyle{IEEEtran}
\bibliography{IEEEabrv,icdeReferences}


% \end{thebibliography}

\end{document}